\def\year{2020}\relax
\documentclass[letterpaper]{article} 
\usepackage{aaai20}  
\usepackage{times}  
\usepackage{helvet} 
\usepackage{courier}  
\usepackage[hyphens]{url}  
\usepackage{graphicx} 
\usepackage{graphicx}  
\usepackage{amsmath}
\usepackage{amsfonts}
\usepackage{multirow}
\usepackage{makecell}
\usepackage{booktabs}

\usepackage{tikz}
\usepackage{subfigure}
\usepackage{pinyin}
\usepackage{ifthen}
\usepackage{pgfplots}
\usepackage{pgfplotstable}

\usetikzlibrary{plotmarks}

\usetikzlibrary{arrows}
\usetikzlibrary{decorations}
\usetikzlibrary{decorations.pathreplacing}
\usetikzlibrary{backgrounds}
\usetikzlibrary{fit}
\usetikzlibrary{positioning}
\usetikzlibrary{calc}
\usetikzlibrary{patterns}
\usetikzlibrary{shadows}
\usetikzlibrary[shapes.multipart]

\newdimen\base
\newdimen\XShift
\newdimen\YShift
\newdimen\LayerSep
\newdimen\TimeSep
\newdimen\vech
\newdimen\vecw

\definecolor{ugreen}{cmyk}{1,0,1,0.498}
\definecolor{myblue}{cmyk}{0.8278,0.3333,0,0.2941}
\definecolor{mygreen}{cmyk}{0.6813,0,0.725,0.3725}
\definecolor{myred}{cmyk}{0,0.8855,0.8767,0.1098}
\definecolor{dblue}{cmyk}{1,0.5487,0,0.5569}

\makeatletter

\newdimen\XCoord
\newdimen\YCoord
\newdimen\TMP
\newcommand*{\ExtractCoordinate}[1]{\path (#1); \pgfgetlastxy{\XCoord}{\YCoord};}%
\newcommand*{\ExtractX}[1]{\path (#1); \pgfgetlastxy{\XCoord}{\TMP};}%
\newcommand*{\ExtractY}[1]{\path (#1); \pgfgetlastxy{\TMP}{\YCoord};}%

\newif\ifpgfrectanglesplitusecustompattern
\def\pgf@lib@sh@rs@list@pattern{none}
\tikzset{
  rectangle split part pattern/.code=\def\pgf@lib@sh@rs@list@pattern{#1}\pgfrectanglesplitusecustompatterntrue,
  rectangle split uses custom pattern/.is if=pgfrectanglesplitusecustompattern,
  rectangle split pattern/.code={\pgfqkeys{/tikz/rectangle split pattern}{#1}},
  rectangle split pattern/.cd,
    pattern color/.code=\edef\tikz@pattern@color{#1},
    pattern/.code=%
      \edef\tikz@temp{#1}%
      \ifx\tikz@temp\tikz@nonetext%
        \tikz@mode@fillfalse
      \else%
        \ifx\tikz@temp\pgfutil@empty%
          \let\tikz@pattern\pgfutil@empty
        \else%
          \def\tikz@pattern{#1}%
        \fi%
        \tikz@mode@filltrue%
      \fi,%
      .unknown/.code=
        \let\tikz@pattern@key\pgfkeyscurrentname
        \pgfutil@ifundefined{pgf@pattern@name@\tikz@pattern@key}{
          \pgfqkeys{/tikz/rectangle split pattern}{pattern color/.try/.expanded=\tikz@pattern@key}%
          \ifpgfkeyssuccess\else
            \pgfkeys{/errors/unknown key={/tikz/\tikz@pattern@key}{#1}}%
          \fi
        }{\pgfqkeys{/tikz/rectangle split pattern}{pattern/.expanded=\tikz@pattern@key}}%
  }
\expandafter\def\csname pgf@sh@bbg@rectangle split\expandafter\expandafter\expandafter\endcsname\expandafter\expandafter\expandafter{\csname pgf@sh@bbg@rectangle split\endcsname
    \ifpgfrectanglesplitusecustompattern%
      \pgf@lib@sh@rs@process@list{\pgf@lib@sh@rs@list@pattern}{\parts}%
      {%
        \ifpgfrectanglesplithorizontal%
          \expandafter\let\expandafter\pgf@lib@sh@rs@fill@bottomleft\expandafter=%
            \csname pgf@anchor@rectangle split@south west\endcsname%
          \pgfmathloop%
            \ifnum\pgfmathcounter>\parts%
            \else%
              \pgf@lib@sh@getalpha\pgf@lib@sh@rs@number{\pgfmathcounter}%
              \ifnum\pgfmathcounter=\parts%
                \expandafter\let\expandafter\pgf@lib@sh@rs@fill@topright\expandafter=%
                  \csname pgf@anchor@rectangle split@north east\endcsname%
              \else%
                \expandafter\let\expandafter\pgf@lib@sh@rs@fill@topright\expandafter=%
                  \csname pgf@anchor@rectangle split@\pgf@lib@sh@rs@number\space split north\endcsname%
              \fi%
              \expandafter\ifx\csname pgf@lib@sh@rs@empty@\pgf@lib@sh@rs@number\endcsname\pgfutil@empty%
              \else%
                \expandafter\ifx\csname pgf@lib@sh@rs@\pgf@lib@sh@rs@number @item\endcsname\pgf@lib@sh@rs@nonetext%
                \else%
                  \pgfextract@process\pgf@lib@sh@rs@fill@bottomleft{%
                    \pgf@lib@sh@rs@fill@bottomleft%
                    \advance\pgf@y by\outerysep\relax%
                    \ifnum\pgfmathcounter=1\relax%
                      \advance\pgf@x by\outerxsep\relax%
                    \fi%
                  }%
                  \pgfextract@process\pgf@lib@sh@rs@fill@topright{%
                    \pgf@lib@sh@rs@fill@topright%
                    \advance\pgf@y by-\outerysep\relax%
                    \ifnum\pgfmathcounter=\parts\relax%
                      \advance\pgf@x by-\outerxsep\relax%
                    \fi%
                  }%
                  \ifnum\pgfmathcounter>1\relax%
                    \begingroup\pgfsetcornersarced{\pgfpointorigin}%
                  \fi%
                  \pgfpathmoveto{%
                    \pgf@lib@sh@rs@fill@topright%
                    \pgf@xa=\pgf@x%
                    \pgf@lib@sh@rs@fill@bottomleft%
                    \pgf@x=\pgf@xa%
                  }%
                  \pgfpathlineto{\pgf@lib@sh@rs@fill@bottomleft}%
                  \pgfpathlineto{%
                    \pgf@lib@sh@rs@fill@bottomleft%
                    \pgf@xa=\pgf@x%
                    \pgf@lib@sh@rs@fill@topright%
                    \pgf@x=\pgf@xa%
                  }%
                  \ifnum\pgfmathcounter>1\relax%
                    \endgroup%
                  \fi%
                  \ifnum\pgfmathcounter<\parts%
                    \begingroup\pgfsetcornersarced{\pgfpointorigin}%
                  \fi%
                  \pgfpathlineto{\pgf@lib@sh@rs@fill@topright}%
                  \pgfpathclose%
                  \ifnum\pgfmathcounter<\parts%
                    \endgroup%
                  \fi%
                  \edef\pgf@temp{\noexpand\pgfqkeys{/tikz/rectangle split pattern}{\csname pgf@lib@sh@rs@\pgf@lib@sh@rs@number @item\endcsname}}%
                  \pgf@temp
                  \iftikz@mode@fill
                    \pgfsetfillpattern{\tikz@pattern}{\tikz@pattern@color}%
                    \pgfusepath{fill}%
                  \else
                    \pgfusepath{}%
                  \fi
                \fi%
              \fi%
              \pgfextract@process\pgf@lib@sh@rs@fill@bottomleft{%
                \pgf@lib@sh@rs@fill@bottomleft%
                \pgf@ya=\pgf@y%
                \pgf@lib@sh@rs@fill@topright%
                \pgf@y=\pgf@ya%
                \advance\pgf@y by-\outerysep%
              }%
          \repeatpgfmathloop%
        \else%
          \expandafter\let\expandafter\pgf@lib@sh@rs@fill@topright\expandafter=%
            \csname pgf@anchor@rectangle split@north east\endcsname%
          \pgfmathloop%
            \ifnum\pgfmathcounter>\parts%
            \else%
              \pgf@lib@sh@getalpha\pgf@lib@sh@rs@number{\pgfmathcounter}%
              \ifnum\pgfmathcounter=\parts%
                \expandafter\let\expandafter\pgf@lib@sh@rs@fill@bottomleft\expandafter=%
                  \csname pgf@anchor@rectangle split@south west\endcsname%
              \else%
                \expandafter\let\expandafter\pgf@lib@sh@rs@fill@bottomleft\expandafter=%
                  \csname pgf@anchor@rectangle split@\pgf@lib@sh@rs@number\space split west\endcsname%
              \fi%
              \expandafter\ifx\csname pgf@lib@sh@rs@empty@\pgf@lib@sh@rs@number\endcsname\pgfutil@empty%
              \else%
                \expandafter\ifx\csname pgf@lib@sh@rs@\pgf@lib@sh@rs@number @item\endcsname\pgf@lib@sh@rs@nonetext%
                \else%
                  \pgfextract@process\pgf@lib@sh@rs@fill@bottomleft{%
                    \pgf@lib@sh@rs@fill@bottomleft%
                    \advance\pgf@x by\outerxsep\relax%
                    \ifnum\parts=1\relax%
                      \advance\pgf@y by\outerysep\relax%
                    \else%
                      \ifnum\pgfmathcounter=\parts
                        \advance\pgf@y by\outerysep\relax%
                      \fi%
                    \fi%
                  }%
                  \pgfextract@process\pgf@lib@sh@rs@fill@topright{%
                    \pgf@lib@sh@rs@fill@topright%
                    \advance\pgf@x by-\outerxsep\relax%
                    \ifnum\parts=1\relax%
                      \advance\pgf@y by-\outerysep\relax%
                    \else%
                      \ifnum\pgfmathcounter=1\relax%
                        \advance\pgf@y by-\outerysep\relax%
                      \fi%
                    \fi%
                  }%
                  \pgfpathmoveto{\pgf@lib@sh@rs@fill@bottomleft}%
                  \ifnum\pgfmathcounter>1\relax%
                    \begingroup\pgfsetcornersarced{\pgfpointorigin}%
                  \fi%
                  \pgfpathlineto{%
                    \pgf@lib@sh@rs@fill@bottomleft%
                    \pgf@xa=\pgf@x%
                    \pgf@lib@sh@rs@fill@topright%
                    \pgf@x=\pgf@xa%
                  }%
                  \pgfpathlineto{\pgf@lib@sh@rs@fill@topright}%
                  \ifnum\pgfmathcounter>1\relax%
                    \endgroup%
                  \fi%
                  \ifnum\pgfmathcounter<\parts%
                    \begingroup\pgfsetcornersarced{\pgfpointorigin}%
                  \fi%
                  \pgfpathlineto{%
                    \pgf@lib@sh@rs@fill@topright%
                    \pgf@xa=\pgf@x%
                    \pgf@lib@sh@rs@fill@bottomleft%
                    \pgf@x=\pgf@xa%
                  }%
                  \pgfpathclose%
                  \ifnum\pgfmathcounter<\parts%
                    \endgroup%
                  \fi%
                  \edef\pgf@temp{\noexpand\pgfqkeys{/tikz/rectangle split pattern}{\csname pgf@lib@sh@rs@\pgf@lib@sh@rs@number @item\endcsname}}%
                  \pgf@temp
                  \iftikz@mode@fill
                    \pgfsetfillpattern{\tikz@pattern}{\tikz@pattern@color}%
                    \pgfusepath{fill}%
                  \else
                    \pgfusepath{}%
                  \fi
                \fi%
              \fi%
              \pgfextract@process\pgf@lib@sh@rs@fill@topright{%
                \pgf@lib@sh@rs@fill@topright%
                \pgf@xa=\pgf@x%
                \pgf@lib@sh@rs@fill@bottomleft%
                \pgf@x=\pgf@xa%
                \advance\pgf@x by\outerxsep\relax%
              }%
          \repeatpgfmathloop%
        \fi%
      }%
    \fi%
}

\newif\ifcuboidshade
\newif\ifcuboidemphedge

\tikzset{
  cuboid/.is family,
  cuboid,
  shiftx/.initial=0,
  shifty/.initial=0,
  dimx/.initial=3,
  dimy/.initial=3,
  dimz/.initial=3,
  scale/.initial=1,
  densityx/.initial=1,
  densityy/.initial=1,
  densityz/.initial=1,
  rotation/.initial=0,
  anglex/.initial=0,
  angley/.initial=90,
  anglez/.initial=225,
  scalex/.initial=1,
  scaley/.initial=1,
  scalez/.initial=0.5,
  front/.style={draw=black,fill=white},
  top/.style={draw=black,fill=white},
  right/.style={draw=black,fill=white},
  shade/.is if=cuboidshade,
  shadecolordark/.initial=black,
  shadecolorlight/.initial=white,
  shadeopacity/.initial=0.15,
  shadesamples/.initial=16,
  emphedge/.is if=cuboidemphedge,
  emphstyle/.style={thick},
}

\newcommand{\tikzcuboidkey}[1]{\pgfkeysvalueof{/tikz/cuboid/#1}}

\newcommand{\tikzcuboid}[1]{
  \tikzset{cuboid,#1} 
  \pgfmathsetlengthmacro{\vectorxx}{\tikzcuboidkey{scalex}*cos(\tikzcuboidkey{anglex})*28.452756}
  \pgfmathsetlengthmacro{\vectorxy}{\tikzcuboidkey{scalex}*sin(\tikzcuboidkey{anglex})*28.452756}
  \pgfmathsetlengthmacro{\vectoryx}{\tikzcuboidkey{scaley}*cos(\tikzcuboidkey{angley})*28.452756}
  \pgfmathsetlengthmacro{\vectoryy}{\tikzcuboidkey{scaley}*sin(\tikzcuboidkey{angley})*28.452756}
  \pgfmathsetlengthmacro{\vectorzx}{\tikzcuboidkey{scalez}*cos(\tikzcuboidkey{anglez})*28.452756}
  \pgfmathsetlengthmacro{\vectorzy}{\tikzcuboidkey{scalez}*sin(\tikzcuboidkey{anglez})*28.452756}
  
  \begin{scope}[xshift=\tikzcuboidkey{shiftx}, yshift=\tikzcuboidkey{shifty}, scale=\tikzcuboidkey{scale}, rotate=\tikzcuboidkey{rotation}, x={(\vectorxx,\vectorxy)}, y={(\vectoryx,\vectoryy)}, z={(\vectorzx,\vectorzy)}]
    \pgfmathsetmacro{\steppingx}{1/\tikzcuboidkey{densityx}}
    \pgfmathsetmacro{\steppingy}{1/\tikzcuboidkey{densityy}}
    \pgfmathsetmacro{\steppingz}{1/\tikzcuboidkey{densityz}}
    \newcommand{\dimx}{\tikzcuboidkey{dimx}}
    \newcommand{\dimy}{\tikzcuboidkey{dimy}}
    \newcommand{\dimz}{\tikzcuboidkey{dimz}}
    \pgfmathsetmacro{\secondx}{2*\steppingx}
    \pgfmathsetmacro{\secondy}{2*\steppingy}
    \pgfmathsetmacro{\secondz}{2*\steppingz}
    \ifthenelse{\equal{\dimx}{1}}
      {\foreach \x in {\steppingx,...,\dimx}}
      {\foreach \x in {\steppingx,\secondx,...,\dimx}}
    {     
      \ifthenelse{\equal{\dimy}{1}}
        {\foreach \y in {\steppingy,...,\dimy}}
        {\foreach \y in {\steppingy,\secondy,...,\dimy}}
      { 
        \pgfmathsetmacro{\lowx}{(\x-\steppingx)}
        \pgfmathsetmacro{\lowy}{(\y-\steppingy)}
        \filldraw[cuboid/front] (\lowx,\lowy,\dimz) -- (\lowx,\y,\dimz) -- (\x,\y,\dimz) -- (\x,\lowy,\dimz) -- cycle;
      }
    }
    \ifthenelse{\equal{\dimx}{1}}
      {\foreach \x in {\steppingx,...,\dimx}}
      {\foreach \x in {\steppingx,\secondx,...,\dimx}}
    { 
      \ifthenelse{\equal{\dimz}{1}}
        {\foreach \z in {\steppingz,...,\dimz}}
        {\foreach \z in {\steppingz,\secondz,...,\dimz}}
      { 
        \pgfmathsetmacro{\lowx}{(\x-\steppingx)}
        \pgfmathsetmacro{\lowz}{(\z-\steppingz)}
        \filldraw[cuboid/top] (\lowx,\dimy,\lowz) -- (\lowx,\dimy,\z) -- (\x,\dimy,\z) -- (\x,\dimy,\lowz) -- cycle;
      }
    }
    \ifthenelse{\equal{\dimy}{1}}
      {\foreach \y in {\steppingy,...,\dimy}}
      {\foreach \y in {\steppingy,\secondy,...,\dimy}}
    { 
      \ifthenelse{\equal{\dimz}{1}}
        {\foreach \z in {\steppingz,...,\dimz}}
        {\foreach \z in {\steppingz,\secondz,...,\dimz}}
      { 
        \pgfmathsetmacro{\lowy}{(\y-\steppingy)}
        \pgfmathsetmacro{\lowz}{(\z-\steppingz)}
        \filldraw[cuboid/right] (\dimx,\lowy,\lowz) -- (\dimx,\lowy,\z) -- (\dimx,\y,\z) -- (\dimx,\y,\lowz) -- cycle;
      }
    }

    \ifcuboidemphedge
      \draw[cuboid/emphstyle] (0,\dimy,0) -- (\dimx,\dimy,0) -- (\dimx,\dimy,\dimz) -- (0,\dimy,\dimz) -- cycle;%
      \draw[cuboid/emphstyle] (0,\dimy,\dimz) -- (0,0,\dimz) -- (\dimx,0,\dimz) -- (\dimx,\dimy,\dimz);%
      \draw[cuboid/emphstyle] (\dimx,\dimy,0) -- (\dimx,0,0) -- (\dimx,0,\dimz);%
    \fi

    \ifcuboidshade
      \pgfmathsetmacro{\cstepx}{\dimx/\tikzcuboidkey{shadesamples}}
      \pgfmathsetmacro{\cstepy}{\dimy/\tikzcuboidkey{shadesamples}}
      \pgfmathsetmacro{\cstepz}{\dimz/\tikzcuboidkey{shadesamples}}
      \foreach \s in {1,...,\tikzcuboidkey{shadesamples}}
      {   
        \pgfmathsetmacro{\lows}{\s-1}
        \pgfmathsetmacro{\cpercent}{(\lows)/(\tikzcuboidkey{shadesamples}-1)*100}
        \fill[opacity=\tikzcuboidkey{shadeopacity},color=\tikzcuboidkey{shadecolorlight}!\cpercent!\tikzcuboidkey{shadecolordark}] (0,\s*\cstepy,\dimz) -- (\s*\cstepx,\s*\cstepy,\dimz) -- (\s*\cstepx,0,\dimz) -- (\lows*\cstepx,0,\dimz) -- (\lows*\cstepx,\lows*\cstepy,\dimz) -- (0,\lows*\cstepy,\dimz) -- cycle;
        \fill[opacity=\tikzcuboidkey{shadeopacity},color=\tikzcuboidkey{shadecolorlight}!\cpercent!\tikzcuboidkey{shadecolordark}] (0,\dimy,\s*\cstepz) -- (\s*\cstepx,\dimy,\s*\cstepz) -- (\s*\cstepx,\dimy,0) -- (\lows*\cstepx,\dimy,0) -- (\lows*\cstepx,\dimy,\lows*\cstepz) -- (0,\dimy,\lows*\cstepz) -- cycle;
        \fill[opacity=\tikzcuboidkey{shadeopacity},color=\tikzcuboidkey{shadecolorlight}!\cpercent!\tikzcuboidkey{shadecolordark}] (\dimx,0,\s*\cstepz) -- (\dimx,\s*\cstepy,\s*\cstepz) -- (\dimx,\s*\cstepy,0) -- (\dimx,\lows*\cstepy,0) -- (\dimx,\lows*\cstepy,\lows*\cstepz) -- (\dimx,0,\lows*\cstepz) -- cycle;
      }
    \fi

  \end{scope}
}

\newcommand{\tikzcuboidface}[4]{
  \begin{scope}[xshift=\tikzcuboidkey{shiftx}, yshift=\tikzcuboidkey{shifty}, scale=\tikzcuboidkey{scale}, rotate=\tikzcuboidkey{rotation}, x={(\vectorxx,\vectorxy)}, y={(\vectoryx,\vectoryy)}, z={(\vectorzx,\vectorzy)}]
    \pgfmathsetmacro{\steppingx}{1/\tikzcuboidkey{densityx}}
    \pgfmathsetmacro{\steppingy}{1/\tikzcuboidkey{densityy}}
    \pgfmathsetmacro{\steppingz}{1/\tikzcuboidkey{densityz}}
    \newcommand{\dimx}{\tikzcuboidkey{dimx}}
    \newcommand{\dimy}{\tikzcuboidkey{dimy}}
    \newcommand{\dimz}{\tikzcuboidkey{dimz}}
    \ifthenelse{\equal{#1}{front}}
      {
        \newcommand{\x}{#3}
        \newcommand{\y}{#4}
        \pgfmathsetmacro{\lowx}{(\x-\steppingx)}
        \pgfmathsetmacro{\lowy}{(\y-\steppingy)}
        \path[#2] (\lowx,\lowy,\dimz) -- (\lowx,\y,\dimz) -- (\x,\y,\dimz) -- (\x,\lowy,\dimz) -- cycle;
      }
      {
        \ifthenelse{\equal{#1}{top}}
          {
            \newcommand{\x}{#3}
            \newcommand{\z}{#4}
            \pgfmathsetmacro{\lowx}{(\x-\steppingx)}
            \pgfmathsetmacro{\lowz}{(\z-\steppingz)}
            \path[#2] (\lowx,\dimy,\lowz) -- (\lowx,\dimy,\z) -- (\x,\dimy,\z) -- (\x,\dimy,\lowz) -- cycle;
          }
          {
            \ifthenelse{\equal{#1}{right}}
              {
                \newcommand{\y}{#3}
                \newcommand{\z}{#4}
                \pgfmathsetmacro{\lowy}{(\y-\steppingy)}
                \pgfmathsetmacro{\lowz}{(\z-\steppingz)}
                \path[#2] (\dimx,\lowy,\lowz) -- (\dimx,\lowy,\z) -- (\dimx,\y,\z) -- (\dimx,\y,\lowz) -- cycle;
              }
              {}
          }
      }
  \end{scope}
}

\newcommand{\tikzcuboidcoordinate}[3]{
  \newdimen\X
  \newdimen\Y
  \newdimen\Z
  \begin{scope}[xshift=\tikzcuboidkey{shiftx}, yshift=\tikzcuboidkey{shifty}, scale=\tikzcuboidkey{scale}, rotate=\tikzcuboidkey{rotation}, x={(\vectorxx,\vectorxy)}, y={(\vectoryx,\vectoryy)}, z={(\vectorzx,\vectorzy)}]
    \pgfmathsetmacro{\steppingx}{1/\tikzcuboidkey{densityx}}
    \pgfmathsetmacro{\steppingy}{1/\tikzcuboidkey{densityy}}
    \pgfmathsetmacro{\steppingz}{1/\tikzcuboidkey{densityz}}
    \newcommand{\dimx}{\tikzcuboidkey{dimx}}
    \newcommand{\dimy}{\tikzcuboidkey{dimy}}
    \newcommand{\dimz}{\tikzcuboidkey{dimz}}
    \ifthenelse{\equal{#1}{front}}
      {\let\Z\dimz}
      {\def\Z{0}}
    \ifthenelse{\equal{#2}{top}}
      {\let\Y\dimy}
      {\def\Y{0}}
    \ifthenelse{\equal{#3}{left}}
      {\def\X{0}}
      {\let\X\dimx}
    
    \coordinate (TMP) at (\X,\Y,\Z);
  \end{scope}
}

\makeatother

\newcommand{\citet}[1]{\citeauthor{#1} \shortcite{#1}}

\title{Neural Machine Translation with Joint Representation}
\author{Yanyang Li\textsuperscript{\rm 1}, Qiang Wang\textsuperscript{\rm 1}, Tong Xiao\textsuperscript{\rm 1,2}\thanks{Corresponding Author.}, Tongran Liu\textsuperscript{\rm 3} and Jingbo Zhu\textsuperscript{\rm 1,2}\\
\textsuperscript{\rm 1}Natural Language Processing Lab., Northeastern University, Shenyang, China\\ 
\textsuperscript{\rm 2}NiuTrans Co., Ltd., Shenyang, China\\
\textsuperscript{\rm 3}CAS Key Laboratory of Behavioral Science, Institute of Psychology, CAS, Beijing, China\\
blamedrlee@outlook.com, wangqiangneu@gmail.com, \{xiaotong, zhujingbo\}@mail.neu.edu.cn, liutr@psych.ac.cn 
}
\begin{document}

  \maketitle

  \begin{abstract}
    Though early successes of Statistical Machine Translation (SMT) systems are attributed in part to the explicit modelling of the interaction between any two source and target units, e.g., alignment, the recent Neural Machine Translation (NMT) systems resort to the attention which partially encodes the interaction for efficiency. In this paper, we employ \emph{Joint Representation} that fully accounts for each possible interaction. We sidestep the inefficiency issue by refining representations with the proposed efficient attention operation. The resulting \emph{Reformer} models offer a new Sequence-to-Sequence modelling paradigm besides the Encoder-Decoder framework and outperform the Transformer baseline in either the small scale IWSLT14 German-English, English-German and IWSLT15 Vietnamese-English or the large scale NIST12 Chinese-English translation tasks by about 1 BLEU point. We also propose a systematic model scaling approach, allowing the Reformer model to beat the state-of-the-art Transformer in IWSLT14 German-English and NIST12 Chinese-English with about 50\% fewer parameters. The code is publicly available at \url{https://github.com/lyy1994/reformer}.
  \end{abstract}

  \section{Introduction}

  To translate one sentence in the source language to its equivalent in the target one, the translation model relies on the bilingual interaction between any two source and target units to select the appropriate hypothesis. Early SMT systems are good examples of this as they use the alignment matrix between the source sentence and the translated part to direct the decoding \cite{book2009:koehn}.

  When it comes to NMT, the natural idea to explicitly model the interaction is by extending the intrinsic representation to have the size $S \times T \times H$, dubbed \emph{Joint Representation}, where $S$ is the source sentence length, $T$ is the target sentence length and $H$ is the hidden size. Despite that this representation can flexibly learn to encode various types of interaction, it is inefficient as it incurs high computation and storage cost. A practical surrogate is the well-known attention mechanism \cite{nips2017:Vaswani}. It mimics the desired interaction by dynamically aggregating a sequence of representations. Though successful, the receptive field of each position in the attention is restricted to one source or target sequence only instead of the cartesian product of them as required by the joint representation.

  In this work, we take one step toward the model family \emph{Reformer}, which built entirely on top of the joint representation. We efficiently adapt the most advanced self-attention module to the joint representation space $\mathbb{R}^{S \times T \times H}$, namely \emph{Separable Attention}. With this building block at hand, we present two instantiations of Reformer. The former one, \emph{Reformer-base}, enjoys the best theoretical effectiveness with the capability that it can access any source or target token with minimum $O(1)$ operations but has higher complexity induced by stacking separable attentions. The latter one, \emph{Reformer-fast}, better trades off the effectiveness and efficiency of the separable attention, achieving comparable results as Reformer-base but 50\% faster. As both Reformer variants do not resort to either the encoder or the decoder, they shed light on exploring the new promising Sequence-to-Sequence paradigm in the future.

  We additionally show that with proper model scaling, our Reformer models are superior to the state-of-the-art (SOTA) Transformer \cite{nips2017:Vaswani} with fewer parameters on larger datasets. The proposed model scaling method requires only $O(2)$ runs to generate the enhanced model compared to the common Grid Search, whose cost grows polynomially as the candidates of hyper-parameters increase.

  In our experiments, the Reformer models achieve 1.3, 0.8 and 0.7 BLEU point improvement over the Transformer baseline in the small scale IWSLT15 Vietnamese-English and IWSLT14 German-English, English-German datasets as well as 1.9 in the large scale NIST12 Chinese-English dataset. After scaling, it outperforms the SOTA large Transformer counterpart by 0.7 and 2 BLEU point with about 50\% parameters in IWSLT14 German-English and NIST12 Chinese-English translations respectively.
  
  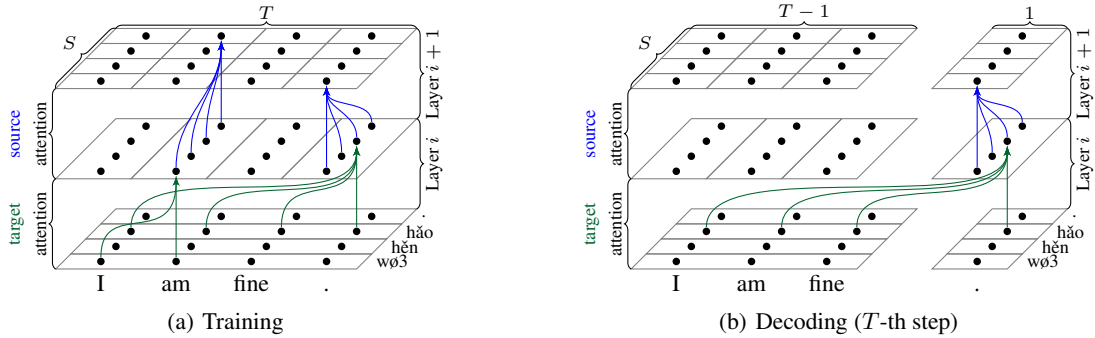
\begin{figure*}[t!]
    \centering
    \hspace*{\fill}
    \subfigure[Training]
    {
      \centering
      \begin{tikzpicture}

        \YShift=0.2\base
        \XShift=\base
        \LayerSep=1.2\base
  
        \tikzstyle{reprnode} = [circle,fill=black,inner sep=0pt,minimum size=0.1\base,anchor=center]
        \tikzstyle{bgnode} = [rectangle,draw=gray,inner sep=0pt,outer sep=0pt,xslant=1,minimum width=\XShift,minimum height=\YShift]
  
        \coordinate (start) at (0,0);
  
        \begin{scope}
          \coordinate (bottom00) at (start);
          \foreach \x / \prevx in {1/0, 2/1, 3/2, 4/3}
            \coordinate (bottom\x0) at ([xshift=\XShift]bottom\prevx0);
          \foreach \y / \prevy / \x in {
            1/0/1, 1/0/2, 1/0/3, 1/0/4,
            2/1/1, 2/1/2, 2/1/3, 2/1/4,
            3/2/1, 3/2/2, 3/2/3, 3/2/4,
            4/3/1, 4/3/2, 4/3/3, 4/3/4}
          {
            \node[reprnode] (bottom\x\y) at ([xshift=\YShift,yshift=\YShift]bottom\x\prevy) {};
          }
        \end{scope}
  
        \foreach \y in {1,2,3,4}
        {
          \node[bgnode,opacity=0] (blockstart) at (bottom1\y) {};
          \node[bgnode,opacity=0] (blockend) at (bottom4\y) {};
          \draw[gray,xslant=1] (blockstart.south west) rectangle (blockend.north east);
        }
  
        \begin{scope}
          \coordinate (middle00) at ([yshift=\LayerSep]bottom00.north);
          \foreach \x / \prevx in {1/0, 2/1, 3/2, 4/3}
            \coordinate (middle\x0) at ([xshift=\XShift]middle\prevx0);
          \foreach \y / \prevy / \x in {
            1/0/1, 1/0/2, 1/0/3, 1/0/4,
            2/1/1, 2/1/2, 2/1/3, 2/1/4,
            3/2/1, 3/2/2, 3/2/3, 3/2/4,
            4/3/1, 4/3/2, 4/3/3, 4/3/4}
          {
            \node[reprnode] (middle\x\y) at ([xshift=\YShift,yshift=\YShift]middle\x\prevy) {};
          }
        \end{scope}
  
        \foreach \x in {1,2,3,4}
        {
          \node[bgnode,opacity=0] (blockstart) at (middle\x1) {};
          \node[bgnode,opacity=0] (blockend) at (middle\x4) {};
          \draw[gray,xslant=1] (blockstart.south west) rectangle (blockend.north east);
        }
  
        \begin{scope}
          \coordinate (top00) at ([yshift=\LayerSep]middle00.north);
          \foreach \x / \prevx in {1/0, 2/1, 3/2, 4/3}
            \coordinate (top\x0) at ([xshift=\XShift]top\prevx0);
          \foreach \y / \prevy / \x in {
            1/0/1, 1/0/2, 1/0/3, 1/0/4,
            2/1/1, 2/1/2, 2/1/3, 2/1/4,
            3/2/1, 3/2/2, 3/2/3, 3/2/4,
            4/3/1, 4/3/2, 4/3/3, 4/3/4}
          {
            \node[reprnode] (top\x\y) at ([xshift=\YShift,yshift=\YShift]top\x\prevy) {};
            \node[bgnode] () at (top\x\y) {};
          }
        \end{scope}
  
        \draw[-latex',thin,ugreen] (bottom13.north) .. controls +(north:\base) and +(south:\base) .. (middle43.south);
        \draw[-latex',thin,ugreen] (bottom23.north) .. controls +(north:0.9\base) and +(south:\base) .. (middle43.south);
        \draw[-latex',thin,ugreen] (bottom33.north) .. controls +(north:0.8\base) and +(south:0.9\base) .. (middle43.south);
        \draw[-latex',thin,ugreen] (bottom43.north) to [out=90,in=-90] (middle43.south);
        
        \draw[-latex',thin,ugreen] (bottom11.north) .. controls +(north:0.8\base) and +(south:\base) .. (middle21.south);
        \draw[-latex',thin,ugreen] (bottom21.north) to [out=90,in=-90] (middle21.south);
  
        \draw[-latex',thin,blue] (middle21.north) to [out=90,in=-90] (top24.south);
        \draw[-latex',thin,blue] (middle22.north) to [out=90,in=-90] (top24.south);
        \draw[-latex',thin,blue] (middle23.north) to [out=90,in=-90] (top24.south);
        \draw[-latex',thin,blue] (middle24.north) to [out=90,in=-90] (top24.south);
        
        \draw[-latex',thin,blue] (middle41.north) to [out=90,in=-90] (top41.south);
        \draw[-latex',thin,blue] (middle42.north) to [out=90,in=-90] (top41.south);
        \draw[-latex',thin,blue] (middle43.north) to [out=90,in=-90] (top41.south);
        \draw[-latex',thin,blue] (middle44.north) to [out=90,in=-90] (top41.south);
  
        \begin{scope}[decoration=brace]
          \node[bgnode,opacity=0] (blockstart) at (middle44) {};
          \node[bgnode,opacity=0] (blockend) at (bottom44) {};
          \draw[decorate] (blockstart.north east) to node [auto,anchor=north,rotate=90,align=center,font=\scriptsize] {Layer $i$} (blockend.north east);

          \node[bgnode,opacity=0] (blockstart) at (top44) {};
          \node[bgnode,opacity=0] (blockend) at (middle44) {};
          \draw[decorate] (blockstart.north east) to node [auto,anchor=north,rotate=90,align=center,font=\scriptsize] {Layer $i+1$} (blockend.north east);
  
          \node[bgnode,opacity=0] (blockstart) at (bottom11) {};
          \node[bgnode,opacity=0] (blockend) at (middle11) {};
          \draw[decorate] (blockstart.south west) to node [auto,anchor=south,rotate=90,align=center,font=\scriptsize] {{\color{ugreen} target}\\[0pt]attention} (blockend.south west);
  
          \node[bgnode,opacity=0] (blockstart) at (middle11) {};
          \node[bgnode,opacity=0] (blockend) at (top11) {};
          \draw[decorate] (blockstart.south west) to node [auto,anchor=south,rotate=90,align=center,font=\scriptsize] {{\color{blue} source}\\[0pt]attention} (blockend.south west);
  
          \node[bgnode,opacity=0] (blockstart) at (top14) {};
          \node[bgnode,opacity=0] (blockend) at (top11) {};
          \draw[decorate,decoration={brace,mirror}] (blockstart.north west) to node [above left] {\scriptsize $S$} (blockend.south west);
  
          \node[bgnode,opacity=0] (blockstart) at (top14) {};
          \node[bgnode,opacity=0] (blockend) at (top44) {};
          \draw[decorate] (blockstart.north west) to node [auto] {\scriptsize $T$} (blockend.north east);
        \end{scope}
  
        \tikzstyle{srcnode} = [inner sep=0pt,anchor=west,font=\scriptsize]
        \node[srcnode] (src1) at ([xshift=0.6\XShift]bottom41.east) {w\o3};
        \node[srcnode] (src2) at ([xshift=0.6\XShift]bottom42.east) {h\en3};
        \node[srcnode] (src3) at ([xshift=0.6\XShift]bottom43.east) {h\ao3};
        \node[srcnode] (src4) at ([xshift=0.6\XShift]bottom44.east) {.};
  
        \tikzstyle{tgtnode} = [font=\footnotesize]
        \node[tgtnode,anchor=north] (tgt1) at (bottom11.south) {I};
        \ExtractX{$(bottom21.south)$};
        \ExtractY{$(tgt1.base)$}; 
        \node[tgtnode,anchor=base] (tgt2) at (\XCoord,\YCoord) {am};
        \ExtractX{$(bottom31.south)$};
        \ExtractY{$(tgt1.base)$};
        \node[tgtnode,anchor=base] (tgt3) at (\XCoord,\YCoord) {fine};
        \ExtractX{$(bottom41.south)$};
        \ExtractY{$(tgt1.base)$};
        \node[tgtnode,anchor=base] (tgt4) at (\XCoord,\YCoord) {.};
        
      \end{tikzpicture}
      \label{fig:train}
    }
    \hfill
    \subfigure[Decoding ($T$-th step)]
    {
      \centering
      \begin{tikzpicture}

        \YShift=0.2\base
        \XShift=\base
        \LayerSep=1.2\base
        \TimeSep=2\base
  
        \tikzstyle{reprnode} = [circle,fill=black,inner sep=0pt,minimum size=0.1\base,anchor=center]
        \tikzstyle{bgnode} = [rectangle,draw=gray,inner sep=0pt,outer sep=0pt,xslant=1,minimum width=\XShift,minimum height=\YShift]
  
        \coordinate (start) at (0,0);
  
        \begin{scope}
          \coordinate (bottom00) at (start);
          \foreach \x / \prevx in {1/0, 2/1, 3/2}
            \coordinate (bottom\x0) at ([xshift=\XShift]bottom\prevx0);
          \coordinate (bottom40) at ([xshift=\TimeSep]bottom30);
          \foreach \y / \prevy / \x in {
            1/0/1, 1/0/2, 1/0/3, 1/0/4,
            2/1/1, 2/1/2, 2/1/3, 2/1/4,
            3/2/1, 3/2/2, 3/2/3, 3/2/4,
            4/3/1, 4/3/2, 4/3/3, 4/3/4}
          {
            \node[reprnode] (bottom\x\y) at ([xshift=\YShift,yshift=\YShift]bottom\x\prevy) {};
          }
        \end{scope}
  
        \foreach \y in {1,2,3,4}
        {
          \node[bgnode,opacity=0] (blockstart) at (bottom1\y) {};
          \node[bgnode,opacity=0] (blockend) at (bottom3\y) {};
          \draw[gray,xslant=1] (blockstart.south west) rectangle (blockend.north east);
          \node[bgnode] () at (bottom4\y) {};
        }
  
        \begin{scope}
          \coordinate (middle00) at ([yshift=\LayerSep]bottom00.north);
          \foreach \x / \prevx in {1/0, 2/1, 3/2}
            \coordinate (middle\x0) at ([xshift=\XShift]middle\prevx0);
          \coordinate (middle40) at ([xshift=\TimeSep]middle30);
          \foreach \y / \prevy / \x in {
            1/0/1, 1/0/2, 1/0/3, 1/0/4,
            2/1/1, 2/1/2, 2/1/3, 2/1/4,
            3/2/1, 3/2/2, 3/2/3, 3/2/4,
            4/3/1, 4/3/2, 4/3/3, 4/3/4}
          {
            \node[reprnode] (middle\x\y) at ([xshift=\YShift,yshift=\YShift]middle\x\prevy) {};
          }
        \end{scope}
  
        \foreach \x in {1,2,3,4}
        {
          \node[bgnode,opacity=0] (blockstart) at (middle\x1) {};
          \node[bgnode,opacity=0] (blockend) at (middle\x4) {};
          \draw[gray,xslant=1] (blockstart.south west) rectangle (blockend.north east);
        }
  
        \begin{scope}
          \coordinate (top00) at ([yshift=\LayerSep]middle00.north);
          \foreach \x / \prevx in {1/0, 2/1, 3/2}
            \coordinate (top\x0) at ([xshift=\XShift]top\prevx0);
          \coordinate (top40) at ([xshift=\TimeSep]top30);
          \foreach \y / \prevy / \x in {
            1/0/1, 1/0/2, 1/0/3, 1/0/4,
            2/1/1, 2/1/2, 2/1/3, 2/1/4,
            3/2/1, 3/2/2, 3/2/3, 3/2/4,
            4/3/1, 4/3/2, 4/3/3, 4/3/4}
          {
            \node[reprnode] (top\x\y) at ([xshift=\YShift,yshift=\YShift]top\x\prevy) {};
            \node[bgnode] () at (top\x\y) {};
          }
        \end{scope}
  
        \draw[-latex',thin,ugreen] (bottom13.north) .. controls +(north:\base) and +(south:\base) .. (middle43.south);
        \draw[-latex',thin,ugreen] (bottom23.north) .. controls +(north:0.9\base) and +(south:\base) .. (middle43.south);
        \draw[-latex',thin,ugreen] (bottom33.north) .. controls +(north:0.8\base) and +(south:\base) .. (middle43.south);
        \draw[-latex',thin,ugreen] (bottom43.north) to [out=90,in=-90] (middle43.south);
  
        \draw[-latex',thin,blue] (middle41.north) to [out=90,in=-90] (top41.south);
        \draw[-latex',thin,blue] (middle42.north) to [out=90,in=-90] (top41.south);
        \draw[-latex',thin,blue] (middle43.north) to [out=90,in=-90] (top41.south);
        \draw[-latex',thin,blue] (middle44.north) to [out=90,in=-90] (top41.south);
  
        \begin{scope}[decoration=brace]
          \node[bgnode,opacity=0] (blockstart) at (middle44) {};
          \node[bgnode,opacity=0] (blockend) at (bottom44) {};
          \draw[decorate] (blockstart.north east) to node [auto,anchor=north,rotate=90,align=center,font=\scriptsize] {Layer $i$} (blockend.north east);

          \node[bgnode,opacity=0] (blockstart) at (top44) {};
          \node[bgnode,opacity=0] (blockend) at (middle44) {};
          \draw[decorate] (blockstart.north east) to node [auto,anchor=north,rotate=90,align=center,font=\scriptsize] {Layer $i+1$} (blockend.north east);
  
          \node[bgnode,opacity=0] (blockstart) at (bottom11) {};
          \node[bgnode,opacity=0] (blockend) at (middle11) {};
          \draw[decorate] (blockstart.south west) to node [auto,anchor=south,rotate=90,align=center,font=\scriptsize] {{\color{ugreen} target}\\[0pt]attention} (blockend.south west);
  
          \node[bgnode,opacity=0] (blockstart) at (middle11) {};
          \node[bgnode,opacity=0] (blockend) at (top11) {};
          \draw[decorate] (blockstart.south west) to node [auto,anchor=south,rotate=90,align=center,font=\scriptsize] {{\color{blue} source}\\[0pt]attention} (blockend.south west);
  
          \node[bgnode,opacity=0] (blockstart) at (top14) {};
          \node[bgnode,opacity=0] (blockend) at (top34) {};
          \draw[decorate] (blockstart.north west) to node [auto] {\scriptsize $T-1$} (blockend.north east);
  
          \node[bgnode,opacity=0] (blockstart) at (top14) {};
          \node[bgnode,opacity=0] (blockend) at (top11) {};
          \draw[decorate,decoration={brace,mirror}] (blockstart.north west) to node [above left] {\scriptsize $S$} (blockend.south west);
  
          \node[bgnode,opacity=0] (blockstart) at (top44) {};
          \node[bgnode,opacity=0] (blockend) at (top44) {};
          \draw[decorate] (blockstart.north west) to node [auto] {\scriptsize $1$} (blockend.north east);
        \end{scope}
  
        \tikzstyle{srcnode} = [inner sep=0pt,anchor=west,font=\scriptsize]
        \node[srcnode] (src1) at ([xshift=0.6\XShift]bottom41.east) {w\o3};
        \node[srcnode] (src2) at ([xshift=0.6\XShift]bottom42.east) {h\en3};
        \node[srcnode] (src3) at ([xshift=0.6\XShift]bottom43.east) {h\ao3};
        \node[srcnode] (src4) at ([xshift=0.6\XShift]bottom44.east) {.};
  
        \tikzstyle{tgtnode} = [font=\footnotesize]
        \node[tgtnode,anchor=north] (tgt1) at (bottom11.south) {I};
        \ExtractX{$(bottom21.south)$};
        \ExtractY{$(tgt1.base)$}; 
        \node[tgtnode,anchor=base] (tgt2) at (\XCoord,\YCoord) {am};
        \ExtractX{$(bottom31.south)$};
        \ExtractY{$(tgt1.base)$};
        \node[tgtnode,anchor=base] (tgt3) at (\XCoord,\YCoord) {fine};
        \ExtractX{$(bottom41.south)$};
        \ExtractY{$(tgt1.base)$};
        \node[tgtnode,anchor=base] (tgt4) at (\XCoord,\YCoord) {.};
        
      \end{tikzpicture}
      \label{fig:decode}
    }
    \hspace*{\fill}
    \caption{Separable Attention over representations (Chinese pinyin-English: \emph{``w\o3 h\en3 h\ao3 .''} $\to$ \emph{``I am fine .''}).}
    \label{fig:separable-attention}
  \end{figure*}
  
  \section{Background}

  \subsection{Sequence-to-Sequence Learning}
  
  Given a sentence pair $(x,y)$, the NMT model learns to maximize its probability $\mathrm{Pr}(y|x)$, which is decomposed into the product of the conditional probability of each target token $\mathrm{Pr}(y|x)=\prod^T_{t=1}\mathrm{Pr}(y_t|y_{<t},x)$, where $y_t$ is the target token at position $t$ and $y_{<t}$ is all target tokens before $t$. Existing Sequence-to-Sequence models for NMT will first use an encoder to map the source sentence $x$ into a sequence of real value vectors $h$, then a decoder to predict $\mathrm{Pr}(y_t|y_{<t},x)$ using $h$ and $y_{<t}$.

  \subsection{Transformer}
  
  The most successful sequence-to-sequence model is Transformer \cite{nips2017:Vaswani}, which consists of a stack of layers. Each layer first utilizes the self-attention to extract information from the whole sentence, then follows a point-wise feed-forward network to provide non-linearity. These two types of building blocks, self-attention and feed-forward network, are both wrapped by the residual connection \cite{cvpr2016:He} to form a sublayer:
  \begin{equation}
    \mathrm{Sublayer}(x)=\mathrm{Block}(\mathrm{LayerNorm}(x))+x
    \label{eqn:residual}
  \end{equation}
  where $x$ is the input representation, $\mathrm{Block}$ is either the self-attention or the feed-forward network and $\mathrm{LayerNorm}$ is the layer normalization \cite{corr2016:Ba}. The self-attention is formulated as:
  \begin{equation}
    \mathrm{Attention}(Q,K,V)=\mathrm{softmax}(\frac{QK^T}{\sqrt{d_\mathrm{m}}})V
    \label{eqn:self-attention}
  \end{equation}
  where $d_\mathrm{m}$ is the dimension of the hidden representation and set as the embedding size. For the self-attention inside the encoder, $Q,K,V \in \mathbb{R}^{S \times d_\mathrm{m}}$, while for the self-attention inside the decoder, $Q,K,V \in \mathbb{R}^{T \times d_\mathrm{m}}$. For the attention that bridges the encoder and decoder, $Q \in \mathbb{R}^{T \times d_\mathrm{m}}$ and $K,V \in \mathbb{R}^{S \times d_\mathrm{m}}$. As for the feed-forward network, it consists of two linear projections with ReLU activation in between:
  \begin{equation}
    \mathrm{FFN}(x)=\max(0,xW_1+b_1)W_2+b_2
    \label{eqn:ffn}
  \end{equation}
  For more details, please refer to \citet{nips2017:Vaswani}.

  \section{The Reformer Family}

  \subsection{Joint Representation}

  The first problem of designing Reformer is to construct the initial joint representation as the model input. For the encoder or the decoder of a standard NMT model, its input is a sized $S \times E$ or $T \times E$ matrix, where $E$ is the token embedding size. It implies that there are $S$ or $T$ tokens with each represented by a sized $E$ embedding. For the joint representation with the size $S \times T \times H$, it could be interpreted in a similar way: there are $S \times T$ tokens combinations where the first token is from the source sentence and the second one is from the target sentence. Each tokens combination is represented by a sized $H$ embedding.

  However, naively assigning a unique embedding to each possible combination is intractable, as it will result in a $V^2 \times H$ embedding matrix, where $V$ is the vocabulary size typically with the scale $10^3 \sim 10^4$. Intuitively, there are only weak connections (e.g., word translation) among tokens from different languages without knowing the context. The independency among tokens can then be safely imposed to factorize their combination. Therefore the combination of each token embedding becomes the embedding of the tokens combination.

  Another issue is to inject the position information to the joint representation for the attention mechanism, as it is unaware of tokens orders. Inspired by the position embedding of Transformer, we similarly use the sinusoidal position embedding to represent the position information of one token and the combination of each token position embedding as the position embedding of tokens combination, because positions are independent of the ones from another axis. Here we choose the simplest addition to combine embedding. Eventually, for the combination of $i$-th token from the source sentence and $j$-th token from the target sentence, the embedding of their combination and its position embedding are:
  \begin{equation}
    \begin{split}
      \mathrm{embed}_{ij}&=\mathrm{embed}_i+\mathrm{embed}_j\\
      \mathrm{pos}_{ij}&=\mathrm{pos}_i+\mathrm{pos}_j
    \end{split}
    \label{eqn:embed}
  \end{equation}
  where $\mathrm{embed}$ is the token/combination embedding, $\mathrm{pos}$ is the position embedding. The $H$ of joint representation equals to $E$ according to Eq. (\ref{eqn:embed}). The Reformer input is the sum of $\mathrm{embed}_{ij}$ and $\mathrm{pos}_{ij}$ then multiplied by $\sqrt{E}$.

  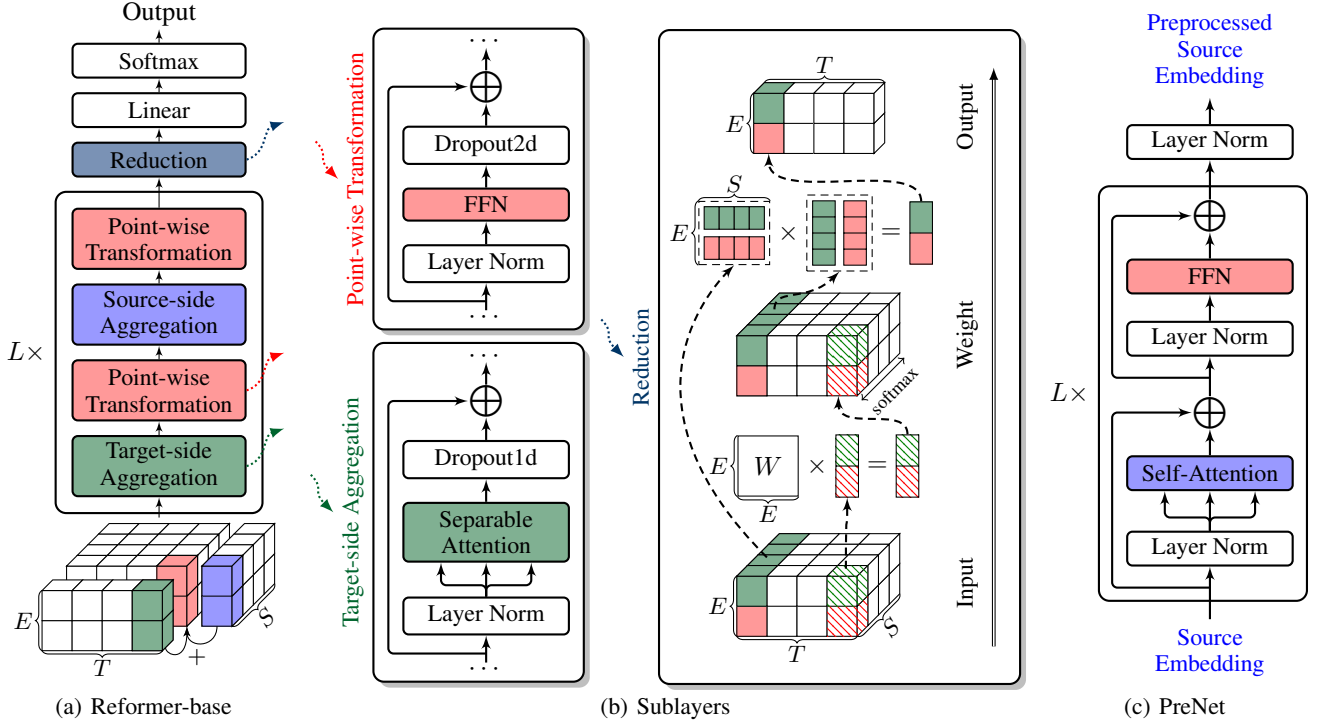
\begin{figure*}[t!]
    \centering
    \hspace*{\fill}
    \subfigure[Reformer-base]
    {
      \centering
      \begin{tikzpicture}
        \LayerSep=0.18\base
  
        \coordinate (start) at (0,0);
  
        \begin{scope}
          \tikzcuboid{%
            shiftx=0,%
            shifty=0,%
            dimx=4,%
            dimy=2,%
            dimz=4,%
            scale=0.4,%
          }
          \tikzcuboidface{front}{draw,fill=red!40}{4}{1}
          \tikzcuboidface{front}{draw,fill=red!40}{4}{2}
          \tikzcuboidface{top}{draw,fill=red!40}{4}{4}
          \tikzcuboidface{right}{draw,fill=red!40}{1}{4}
          \tikzcuboidface{right}{draw,fill=red!40}{2}{4}
    
          \tikzcuboidcoordinate{front}{bottom}{right}
          \coordinate (mix) at (TMP);
    
          \tikzcuboid{%
            shiftx=-0.75\base,%
            shifty=-0.75\base,%
            dimx=4,%
            dimy=2,%
            dimz=1,%
            scale=0.4,%
          }
          \tikzcuboidface{front}{draw,fill=ugreen!40}{4}{1}
          \tikzcuboidface{front}{draw,fill=ugreen!40}{4}{2}
          \tikzcuboidface{top}{draw,fill=ugreen!40}{4}{1}
          \tikzcuboidface{right}{draw,fill=ugreen!40}{1}{1}
          \tikzcuboidface{right}{draw,fill=ugreen!40}{2}{1}
    
          \tikzcuboidcoordinate{front}{bottom}{left}
          \ExtractCoordinate{$(TMP)$}
          \tikzcuboidcoordinate{front}{bottom}{right}
          \draw[decorate,decoration={brace,mirror}] (\XCoord,\YCoord) to node [below] {\small $T$} (TMP);
          \tikzcuboidcoordinate{front}{bottom}{left}
          \ExtractCoordinate{$(TMP)$}
          \tikzcuboidcoordinate{front}{top}{left}
          \draw[decorate,decoration={brace}] (\XCoord,\YCoord) to node [left,midway] {\small $E$} (TMP);
    
          \tikzcuboidcoordinate{back}{bottom}{right}
          \coordinate (tgt) at (TMP);
    
          \tikzcuboid{%
            shiftx=1.8\base,%
            shifty=0,%
            dimx=1,%
            dimy=2,%
            dimz=4,%
            scale=0.4,%
          }
          \tikzcuboidface{front}{draw,fill=blue!40}{1}{1}
          \tikzcuboidface{front}{draw,fill=blue!40}{1}{2}
          \tikzcuboidface{top}{draw,fill=blue!40}{1}{4}
          \tikzcuboidface{right}{draw,fill=blue!40}{1}{4}
          \tikzcuboidface{right}{draw,fill=blue!40}{2}{4}
    
          \tikzcuboidcoordinate{front}{bottom}{right}
          \ExtractCoordinate{$(TMP)$}
          \tikzcuboidcoordinate{back}{bottom}{right}
          \draw[decorate,decoration={brace,mirror}] (\XCoord,\YCoord) to node [rotate=45,below,midway] {\small $S$} (TMP);
    
          \tikzcuboidcoordinate{front}{bottom}{left}
          \coordinate (src) at (TMP);
    
          \tikzcuboidcoordinate{back}{bottom}{right}
          \coordinate (pre) at (TMP);
    
          \draw[-latex'] ([shift={(-0.1\base,-0.1\base)}]tgt) .. controls +(south:0.1\base) and +(south:0.5\base) .. (mix);
          \draw[-latex'] ([xshift=0.2\base]src) .. controls +(south:0.3\base) and +(south:0.3\base) .. (mix);
          \node[font=\small,inner sep=0pt,anchor=north west] () at ([yshift=-0.3\base]mix) {$+$};
    
          \tikzstyle{layernode} = [rectangle,draw,thick,rounded corners=2pt,minimum width=2.3\base,minimum height=0.8\base,font=\small,align=center,inner sep=1pt]
    
          \coordinate (layerstart) at ([shift={(-0.5\base,1.15\base)}]start);
    
          \node[layernode,fill=ugreen!40,anchor=south west] (tgtattn) at (layerstart) {Target-side\\Aggregation};
          \node[layernode,fill=red!40,anchor=south] (tgtffn) at ([yshift=\LayerSep]tgtattn.north) {Point-wise\\Transformation};
          \node[layernode,fill=blue!40,anchor=south] (srcattn) at ([yshift=\LayerSep]tgtffn.north) {Source-side\\Aggregation};
          \node[layernode,fill=red!40,anchor=south] (srcffn) at ([yshift=\LayerSep]srcattn.north) {Point-wise\\Transformation};
    
          \begin{pgfonlayer}{background}
            \coordinate (layerbottomleft) at (tgtattn.south west);
            \coordinate (layertopright) at (srcffn.north east);
            \node[rectangle,thick,draw,inner sep=6pt,rounded corners,fit=(layerbottomleft) (layertopright)] (layer) {};
            \node[left=0pt of layer] () {$L\times$};
          \end{pgfonlayer}
    
          \node[layernode,fill=dblue!40,anchor=south,minimum height=0.45\base] (reduction) at ([yshift=5pt]layer.north) {Reduction};
          \node[layernode,anchor=south,minimum height=0.45\base] (linear) at ([yshift=\LayerSep]reduction.north) {Linear};
          \node[layernode,anchor=south,minimum height=0.45\base] (softmax) at ([yshift=\LayerSep]linear.north) {Softmax};
          \node[above=\LayerSep of softmax,font=\normalsize,inner sep=1pt] (output) {Output};
    
          \draw[-latex'] ([yshift=-1.5\LayerSep]tgtattn.south) to (tgtattn.south);
          \draw[-latex'] (tgtattn.north) to (tgtffn.south);
          \draw[-latex'] (tgtffn.north) to (srcattn.south);
          \draw[-latex'] (srcattn.north) to (srcffn.south);
          \draw[-latex'] (srcffn.north) to (reduction.south);
          \draw[-latex'] (reduction.north) to (linear.south);
          \draw[-latex'] (linear.north) to (softmax.south);
          \draw[-latex'] (softmax.north) to (output.south);
  
          \draw[-latex,thick,densely dotted,ugreen] (tgtattn.east) .. controls +(0.3\base,0.1\base) and +(-0.3\base,-0.1\base) .. ([shift={(0.5\base,0.5\base)}]tgtattn.east);
          \draw[-latex,thick,densely dotted,red] (tgtffn.east) .. controls +(0.3\base,0.1\base) and +(-0.3\base,-0.1\base) .. ([shift={(0.5\base,0.5\base)}]tgtffn.east);
          \draw[-latex,thick,densely dotted,dblue] (reduction.east) .. controls +(0.3\base,0.1\base) and +(-0.3\base,-0.1\base) .. ([shift={(0.5\base,0.5\base)}]reduction.east);
    
        \end{scope}
      \end{tikzpicture}
      \label{fig:reformer-base}
    }
    \hfill
    \subfigure[Sublayers]
    {
      \centering
      \begin{tikzpicture}
        \LayerSep=0.3\base
        
        \begin{scope}[local bounding box=AGGREGATION]
          \coordinate (attnstart) at (0,0);
    
          \tikzstyle{layernode} = [rectangle,draw,thick,rounded corners=2pt,minimum width=2.2\base,font=\small,align=center,inner sep=3pt]
    
          \node[inner sep=0pt,font=\normalsize] (input) at (attnstart) {$\cdots$};
          \node[layernode,above=1.2\LayerSep of input] (norm) {Layer Norm};
          \node[layernode,fill=ugreen!40,above=1.5\LayerSep of norm] (core) {Separable\\Attention};
          \node[layernode,above=\LayerSep of core] (dropout) {Dropout1d};
          \node[inner sep=0pt,font=\Large,above=\LayerSep of dropout] (residual) {$\bigoplus$};
          \node[inner sep=0pt,font=\normalsize,above=\LayerSep of residual] (output) {$\cdots$};
    
          \begin{pgfonlayer}{background}
            \ExtractX{$([xshift=-0.3\base]norm.west)$};
            \ExtractY{$(input.south)$};
            \coordinate (layerbottomleft) at (\XCoord,\YCoord);
            \ExtractX{$([xshift=0.1\base]dropout.east)$};
            \ExtractY{$(output.north)$};
            \coordinate (layertopright) at (\XCoord,\YCoord);
            \node[rectangle,thick,draw,inner sep=1pt,rounded corners,fit=(layerbottomleft) (layertopright)] (corelayer) {};
          \end{pgfonlayer}
    
          \draw[-latex',thick] (input.north) to (norm.south);
          \draw[-latex',thick] (norm.north) to (core.south);
          \draw[-latex',thick] (core.north) to (dropout.south);
          \draw[-latex',thick] (dropout.north) to (residual.south);
          \draw[-latex',thick] (residual.north) to (output.south);
    
          \ExtractX{$([xshift=-0.6\base]core.south)$};
          \ExtractY{$([yshift=0.5\LayerSep]norm.north)$};
          \draw[-latex',thick,rounded corners=2pt] (norm.north) to ([yshift=0.5\LayerSep]norm.north) to (\XCoord,\YCoord) to ([xshift=-0.6\base]core.south);
          \ExtractX{$([xshift=0.6\base]core.south)$};
          \ExtractY{$([yshift=0.5\LayerSep]norm.north)$};
          \draw[-latex',thick,rounded corners=2pt] (norm.north) to ([yshift=0.5\LayerSep]norm.north) to (\XCoord,\YCoord) to ([xshift=0.6\base]core.south);
          \ExtractX{$([shift={(-1.3\base,-0.8\LayerSep)}]norm.south)$};
          \ExtractY{$(residual.west)$};
          \draw[-latex',thick,rounded corners=2pt] ([yshift=-0.8\LayerSep]norm.south) to ([shift={(-1.3\base,-0.8\LayerSep)}]norm.south) to (\XCoord,\YCoord) to (residual.west);
    
          \begin{pgfonlayer}{background}
            \node[rectangle,thick,draw,fill=white,inner sep=1pt,rounded corners,drop shadow,fit=(input) (corelayer) (output)] (attnlayer) {};
          \end{pgfonlayer}
    
        \end{scope}

        \node[font=\small,text=ugreen,outer sep=0pt,left=0pt of attnlayer] (attnlabel) {\rotatebox{90}{Target-side Aggregation}};
        \draw[-latex,thick,densely dotted,ugreen] ([shift={(-0.3\base,0.5\base)}]attnlabel.west) .. controls +(0.2\base,-0.2\base) and +(-0.3\base,0.3\base) .. (attnlabel.west);

        \begin{scope}[local bounding box=FUNCTION]
          \ExtractX{$(attnstart)$}
          \ExtractY{$(AGGREGATION.north)$}
          \coordinate (ffnstart) at ([yshift=0.35\base]\XCoord,\YCoord);
    
          \tikzstyle{layernode} = [rectangle,draw,thick,rounded corners=2pt,minimum width=2.2\base,font=\small,align=center,inner sep=3pt]
    
          \node[inner sep=0pt,font=\normalsize] (input) at (ffnstart) {$\cdots$};
          \node[layernode,above=1.2\LayerSep of input] (norm) {Layer Norm};
          \node[layernode,fill=red!40,above=\LayerSep of norm] (core) {FFN};
          \node[layernode,above=\LayerSep of core] (dropout) {Dropout2d};
          \node[inner sep=0pt,font=\Large,above=\LayerSep of dropout] (residual) {$\bigoplus$};
          \node[inner sep=0pt,font=\normalsize,above=\LayerSep of residual] (output) {$\cdots$};
    
          \begin{pgfonlayer}{background}
            \ExtractX{$([xshift=-0.3\base]norm.west)$};
            \ExtractY{$(input.south)$};
            \coordinate (layerbottomleft) at (\XCoord,\YCoord);
            \ExtractX{$([xshift=0.1\base]dropout.east)$};
            \ExtractY{$(output.north)$};
            \coordinate (layertopright) at (\XCoord,\YCoord);
            \node[rectangle,thick,draw,inner sep=1pt,rounded corners,fit=(layerbottomleft) (layertopright)] (corelayer) {};
          \end{pgfonlayer}
    
          \draw[-latex',thick] (input.north) to (norm.south);
          \draw[-latex',thick] (norm.north) to (core.south);
          \draw[-latex',thick] (core.north) to (dropout.south);
          \draw[-latex',thick] (dropout.north) to (residual.south);
          \draw[-latex',thick] (residual.north) to (output.south);
    
          \ExtractX{$([shift={(-1.3\base,-0.8\LayerSep)}]norm.south)$};
          \ExtractY{$(residual.west)$};
          \draw[-latex',thick,rounded corners=2pt] ([yshift=-0.8\LayerSep]norm.south) to ([shift={(-1.3\base,-0.8\LayerSep)}]norm.south) to (\XCoord,\YCoord) to (residual.west);
    
          \begin{pgfonlayer}{background}
            \node[rectangle,thick,draw,fill=white,inner sep=1pt,rounded corners,drop shadow,fit=(input) (corelayer) (output)] (ffnlayer) {};
          \end{pgfonlayer}
    
        \end{scope}

        \node[font=\small,text=red,outer sep=0pt,left=0pt of ffnlayer] (ffnlabel) {\rotatebox{90}{Point-wise Transformation}};
        \draw[-latex,thick,densely dotted,red] ([shift={(-0.3\base,0.5\base)}]ffnlabel.west) .. controls +(0.2\base,-0.2\base) and +(-0.3\base,0.3\base) .. (ffnlabel.west);
    
        \begin{scope}[local bounding box=REDUCTION]
          \ExtractX{$(AGGREGATION.east)$}
          \ExtractY{$(attnstart.south)$}
          \coordinate (reducestart) at ([shift={(2.6\base,\base)}]\XCoord,\YCoord);
    
          \coordinate (bottomleft) at ([shift={(-1.6\base,-1.2\base)}]reducestart);
          \coordinate (topright) at ([shift={(3.2\base,7.5\base)}]reducestart);
          \begin{pgfonlayer}{background}
            \node[rectangle,thick,draw,fill=white,inner sep=0pt,rounded corners,drop shadow,fit=(bottomleft) (topright)] (reduce) {};
          \end{pgfonlayer}
    
          \ExtractCoordinate{$(reducestart)$};
          \tikzcuboid{%
            shiftx=\XCoord,%
            shifty=\YCoord,%
            dimx=4,%
            dimy=2,%
            dimz=4,%
            scale=0.4,%
          }
          \tikzcuboidface{front}{pattern=north west lines,pattern color=red}{4}{1}
          \tikzcuboidface{front}{pattern=north west lines,pattern color=ugreen}{4}{2}
          \tikzcuboidface{top}{pattern=north west lines,pattern color=ugreen}{4}{4}
          \tikzcuboidface{right}{pattern=north west lines,pattern color=red}{1}{4}
          \tikzcuboidface{right}{pattern=north west lines,pattern color=ugreen}{2}{4}

          \tikzcuboidface{front}{draw,fill=red!40}{1}{1}
          \tikzcuboidface{front}{draw,fill=ugreen!40}{1}{2}
          \tikzcuboidface{top}{draw,fill=ugreen!40}{1}{1}
          \tikzcuboidface{top}{draw,fill=ugreen!40}{1}{2}
          \tikzcuboidface{top}{draw,fill=ugreen!40}{1}{3}
          \tikzcuboidface{top}{draw,fill=ugreen!40}{1}{4}
          \tikzcuboidcoordinate{front}{top}{left}
          \coordinate (xpart) at ([shift={(0.4\base,0.25\base)}]TMP);
          \tikzcuboidcoordinate{front}{top}{right}
          
          \vech=0.8\base
          \vecw=0.3\base
    
          \tikzstyle{vecnode} = [rectangle,draw,inner sep=0pt]
          \tikzstyle{symnode} = [inner sep=0pt,font=\normalsize]
    
          \node[vecnode,minimum width=\vech,minimum height=\vech] (w) at ([xshift=-0.15\base,yshift=1.7\base]reducestart) {$W$};
          \node[symnode,right=0.1\base of w] (matmul) {$\times$};
          \pgfmathparse{\vech/2}
          \node[vecnode,rectangle split,rectangle split parts=2,minimum width=\vecw,rectangle split empty part height=\pgfmathresult,right=0.1\base of matmul,rectangle split part pattern={{north west lines, ugreen},{north west lines, red}}] (repr) {};
          \node[symnode,right=0.1\base of repr] (eq) {$=$};
          \node[vecnode,,rectangle split,rectangle split parts=2,minimum width=\vecw,rectangle split empty part height=\pgfmathresult,right=0.1\base of eq,rectangle split part pattern={{north west lines, ugreen},{north west lines, red}}] (aw) {};
    
          \draw[decorate,decoration=brace] (w.south west) to node [left,midway] {\small $E$} (w.north west);
          \draw[decorate,decoration={brace,mirror}] (w.south west) to node [below,midway] {\small $E$} (w.south east);
          \draw[-latex',thick,densely dashed] ([shift={(-0.15\base,0.1\base)}]TMP.north) .. controls +(north:\base) and +(south:0.6\base) .. (repr.south);
          
          \tikzcuboidcoordinate{front}{bottom}{left}
          \coordinate (TMP2) at (TMP); 
          \tikzcuboidcoordinate{front}{top}{left}
          \draw[decorate,decoration={brace}] (TMP2) to node [left,midway] {\small $E$} (TMP);
          \tikzcuboidcoordinate{front}{bottom}{left}
          \coordinate (TMP2) at (TMP);
          \tikzcuboidcoordinate{front}{bottom}{right}
          \draw[decorate,decoration={brace,mirror}] (TMP2) to node [below,midway] {\small $T$} (TMP);
          \tikzcuboidcoordinate{front}{bottom}{right}
          \coordinate (TMP2) at (TMP);
          \tikzcuboidcoordinate{back}{bottom}{right}
          \draw[decorate,decoration={brace,mirror}] (TMP2) to node [rotate=45,below,midway] {\small $S$} (TMP);
    
          \ExtractCoordinate{$([yshift=3.2\base]reducestart)$};
          \tikzcuboid{%
            shiftx=\XCoord,%
            shifty=\YCoord,%
            dimx=4,%
            dimy=2,%
            dimz=4,%
            scale=0.4,%
          }
          \tikzcuboidface{front}{pattern=north west lines,pattern color=red}{4}{1}
          \tikzcuboidface{front}{pattern=north west lines,pattern color=ugreen}{4}{2}
          \tikzcuboidface{top}{pattern=north west lines,pattern color=ugreen}{4}{4}
          \tikzcuboidface{right}{pattern=north west lines,pattern color=red}{1}{4}
          \tikzcuboidface{right}{pattern=north west lines,pattern color=ugreen}{2}{4}
          \tikzcuboidface{front}{draw,fill=red!40}{1}{1}
          \tikzcuboidface{front}{draw,fill=ugreen!40}{1}{2}
          \tikzcuboidface{top}{draw,fill=ugreen!40}{1}{1}
          \tikzcuboidface{top}{draw,fill=ugreen!40}{1}{2}
          \tikzcuboidface{top}{draw,fill=ugreen!40}{1}{3}
          \tikzcuboidface{top}{draw,fill=ugreen!40}{1}{4}
          \tikzcuboidcoordinate{front}{bottom}{right}
    
          \draw[-latex',thick,densely dashed] (aw.north) .. controls +(north:0.4\base) and +(south:0.5\base) .. ([shift={(-0.25\base,0)}]TMP.south);
    
          \tikzcuboidcoordinate{front}{bottom}{right}
          \coordinate (TMP2) at (TMP);
          \tikzcuboidcoordinate{back}{bottom}{right}
          \draw[<->] ([shift={(0.06\base,-0.06\base)}]TMP2) to node [rotate=45,below,midway] {\scriptsize softmax} ([shift={(0.06\base,-0.06\base)}]TMP);
    
          \pgfmathparse{\vech/4}
    
          \coordinate (x00) at ([yshift=4.2\base,xshift=-\base]reducestart);
          \foreach \curr / \prev in {1/0,2/1}
            \coordinate (x\curr0) at ([yshift=0.1\base+\vecw]x\prev0);
          \foreach \i / \c in {1/red!40,2/ugreen!40}
          {
            \foreach \curr / \prev in {1/0,2/1,3/2,4/3}
            {
              \node[vecnode,minimum width=0.2\base,minimum height=\vecw,inner sep=0pt,outer sep=0pt,fill=\c,right=0pt of x\i\prev] (x\i\curr) {};
            }
          }
          \begin{pgfonlayer}{background}
            \node[rectangle,draw,densely dashed,fill=white,inner sep=2pt,fit=(x11) (x24)] (xmat) {};
          \end{pgfonlayer}
          \draw[decorate,decoration=brace] (xmat.south west) to node [left,midway] {\small $E$} (xmat.north west);
          \draw[decorate,decoration=brace] (xmat.north west) to node [above,midway] {\small $S$} (xmat.north east);
    
          \node[symnode,right=0.1\base of xmat] (matmul) {$\times$};
          \coordinate (aw0) at ([xshift=2pt]matmul.east);
          \foreach \curr / \prev / \c in {1/0/ugreen!40,2/1/red!40}
          {
            \node[vecnode,rectangle split,rectangle split parts=4,minimum width=\vecw,rectangle split empty part height=\pgfmathresult,fill=\c,anchor=west] (aw\curr) at ([xshift=0.1\base]aw\prev.east) {};
          }
          \begin{pgfonlayer}{background}
            \node[rectangle,draw,densely dashed,fill=white,inner sep=2pt,fit=(aw1) (aw2)] (awmat) {};
          \end{pgfonlayer}
    
          \node[symnode,right=0.1\base of awmat] (eq) {$=$};
          \pgfmathparse{\vech / 2}
          \node[vecnode,rectangle split,rectangle split parts=2,minimum width=\vecw,rectangle split empty part height=\pgfmathresult,right=0.1\base of eq,rectangle split part fill={ugreen!40,red!40}] (o) {};
          
          \tikzcuboidcoordinate{front}{top}{left}
          \coordinate (awpart) at ([shift={(0.5\base,0.35\base)}]TMP);
          \draw[-latex',thick,densely dashed] (xpart) .. controls +(-2\base,2\base) and +(south west:0.6\base) .. (xmat.south);
          \draw[-latex',thick,densely dashed] (awpart) .. controls +(north:0.3\base) and +(south west:0.6\base) .. (awmat.south);
    
          \ExtractCoordinate{$([yshift=6\base,xshift=-0.2\base]reducestart)$};
          \tikzcuboid{%
            shiftx=\XCoord,%
            shifty=\YCoord,%
            dimx=4,%
            dimy=2,%
            dimz=1,%
            scale=0.4,%
          }
          \tikzcuboidface{front}{draw,fill=red!40}{1}{1}
          \tikzcuboidface{front}{draw,fill=ugreen!40}{1}{2}
          \tikzcuboidface{top}{draw,fill=ugreen!40}{1}{1}
          \tikzcuboidcoordinate{front}{bottom}{left}
          \draw[-latex',thick,densely dashed] (o.north) .. controls +(north:0.6\base) and +(south:0.7\base) .. ([xshift=0.2\base]TMP.south);

          \tikzcuboidcoordinate{back}{top}{left}
          \coordinate (TMP2) at (TMP);
          \tikzcuboidcoordinate{back}{top}{right}
          \draw[decorate,decoration={brace,mirror}] (TMP) to node [above,midway] {\small $T$} (TMP2);

          \tikzcuboidcoordinate{front}{top}{left}
          \coordinate (TMP2) at (TMP);
          \tikzcuboidcoordinate{front}{bottom}{left}
          \draw[decorate,decoration={brace,mirror}] (TMP2) to node [left,midway] {\small $E$} (TMP);
    
        \end{scope}
    
        \node[font=\small,text=dblue,outer sep=0pt,left=0pt of reduce] (reducelabel) {\rotatebox{90}{Reduction}};
        \draw[-latex,thick,densely dotted,dblue] ([shift={(-0.3\base,0.5\base)}]reducelabel.west) .. controls +(0.2\base,-0.2\base) and +(-0.3\base,0.3\base) .. (reducelabel.west);
    
        \node[anchor=south,rotate=90,font=\footnotesize] () at ([xshift=-0.45\base,yshift=-3\base]reduce.east) {Input};
        \node[anchor=south,rotate=90,font=\footnotesize] () at ([xshift=-0.45\base,yshift=0.3\base]reduce.east) {Weight};
        \node[anchor=south,rotate=90,font=\footnotesize] () at ([xshift=-0.45\base,yshift=3.2\base]reduce.east) {Output};
    
        \ExtractX{$(topright)$}
        \ExtractY{$(bottomleft)$}
        \coordinate (btstart) at ([xshift=-0.35\base,yshift=0.5\base]\XCoord,\YCoord);
        \coordinate (btend) at ([xshift=-0.35\base,yshift=-0.5\base]topright);
        \draw[double,-latex] (btstart) to (btend);
      \end{tikzpicture}
      \label{fig:sublayers}
    }
    \hfill
    \subfigure[PreNet]
    {
      \centering
      \begin{tikzpicture}
        \LayerSep=0.35\base

        \begin{scope}
          \coordinate (prestart) at (0,0);
    
          \tikzstyle{layernode} = [rectangle,draw,thick,rounded corners=2pt,minimum width=2.2\base,minimum height=0.45\base,font=\small,align=center,inner sep=1pt]
    
          \node[text=blue,font=\small,align=center] (input) at (prestart) {Source\\Embedding};
          \node[layernode,above=2\LayerSep of input] (norm1) {Layer Norm};
          \node[layernode,fill=blue!40,above=1.5\LayerSep of norm1] (attn) {Self-Attention};
          \node[inner sep=0pt,font=\Large,above=\LayerSep of attn] (residual1) {$\bigoplus$};
          \node[layernode,above=1.5\LayerSep of residual1] (norm2) {Layer Norm};
          \node[layernode,fill=red!40,above=\LayerSep of norm2] (ffn) {FFN};
          \node[inner sep=0pt,font=\Large,above=\LayerSep of ffn] (residual2) {$\bigoplus$};
          \node[layernode,above=1.5\LayerSep of residual2] (norm3) {Layer Norm};
          \node[text=blue,font=\small,align=center,above=\LayerSep of norm3] (output) {Preprocessed\\Source\\Embedding};
    
          \draw[-latex',thick] (input.north) to (norm1.south);
          \draw[-latex',thick] (norm1.north) to (attn.south);
          \draw[-latex',thick] (attn.north) to (residual1.south);
          \draw[-latex',thick] (residual1.north) to (norm2.south);
          \draw[-latex',thick] (norm2.north) to (ffn.south);
          \draw[-latex',thick] (ffn.north) to (residual2.south);
          \draw[-latex',thick] (residual2.north) to (norm3.south);
          \draw[-latex',thick] (norm3.north) to (output.south);
    
          \ExtractX{$([xshift=-0.6\base]attn.south)$};
          \ExtractY{$([yshift=0.5\LayerSep]norm1.north)$};
          \draw[-latex',thick,rounded corners=2pt] (norm1.north) to ([yshift=0.5\LayerSep]norm1.north) to (\XCoord,\YCoord) to ([xshift=-0.6\base]attn.south);
          \ExtractX{$([xshift=0.6\base]attn.south)$};
          \ExtractY{$([yshift=0.5\LayerSep]norm1.north)$};
          \draw[-latex',thick,rounded corners=2pt] (norm1.north) to ([yshift=0.5\LayerSep]norm1.north) to (\XCoord,\YCoord) to ([xshift=0.6\base]attn.south);
          \ExtractX{$([shift={(-1.3\base,-0.8\LayerSep)}]norm1.south)$};
          \ExtractY{$(residual1.west)$};
          \draw[-latex',thick,rounded corners=2pt] ([yshift=-0.8\LayerSep]norm1.south) to ([shift={(-1.3\base,-0.8\LayerSep)}]norm1.south) to (\XCoord,\YCoord) to (residual1.west);
          \ExtractX{$([shift={(-1.3\base,-0.8\LayerSep)}]norm2.south)$};
          \ExtractY{$(residual2.west)$};
          \draw[-latex',thick,rounded corners=2pt] ([yshift=-0.8\LayerSep]norm2.south) to ([shift={(-1.3\base,-0.8\LayerSep)}]norm2.south) to (\XCoord,\YCoord) to (residual2.west);
    
          \begin{pgfonlayer}{background}
            \coordinate (bottomleft) at ([shift={(-1.3\base,-0.8\LayerSep)}]norm1.south);
            \ExtractX{$(ffn.east)$};
            \ExtractY{$(residual2.north)$};
            \coordinate (topright) at (\XCoord,\YCoord);
            \node[rectangle,thick,draw,fill=white,inner sep=5pt,rounded corners,fit=(bottomleft) (topright)] (preprocesslayer) {};
            \node[left=0pt of preprocesslayer] (label) {$L\times$};
          \end{pgfonlayer}

        \end{scope}
      \end{tikzpicture}
      \label{fig:preprocess}
    }
    \hspace*{\fill}
    \caption{The Reformer Architecture. The green cube in the bottom of Figure. \ref{fig:reformer-base} is the target token embedding and the blue one is the source token embedding if in Reformer-base otherwise the preprocessed source token embedding from Figure. \ref{fig:preprocess}.}
    \label{fig:architecture}
  \end{figure*}

  \subsection{Separable Attention}

  The simplest approach to adapt the self-attention to the $\mathbb{R}^{S \times T \times E}$ input space is to collapse the $S$ and $T$ dimensions into a single dimension $J=S \times T$, then perform self-attention to aggregate different combinations information by treating it as a length $J$ sequence. But such an approach suffers from great inefficiency though it can access any combination with $O(1)$ attention, as the operations required for the self-attention grow quadratically with the sequence length, i.e., $J^2=S^2 \times T^2$ operations.

  Inspired by Separable Convolution \cite{corr2017:Howard} and the fact that combinations are strongly connected along either the $S$ or $T$ dimension, we instead perform the self-attention on these two temporal dimensions separately and sequentially, dubbed \emph{Separable Attention}. Without the lack of theoretical effectiveness, the network output can access the input information in any source or target token through $O(2)$ attentions, as long as we stack one attention operated on $S$ and another one on $T$ alternatively. This way reduces the complexity from $S^2 \times T^2$ to $S^2 \times T + S \times T^2$. Formally, for the separable attention that aggregates information along the dimension $D \in \{S,T\}$, we have:
  \begin{equation}
    \begin{split}
      \mathrm{SepAttn}(Q,K,V)&=\mathrm{Concat}(\mathrm{split}_1,\cdots,\mathrm{split}_D)\\
      \mathrm{where}\ \mathrm{split}_i&=\mathrm{Attention}(Q_i,K_i,V_i)
    \end{split}
    \label{eqn:separable-attention}
  \end{equation}
  where $\mathrm{Attention}$ is the self-attention from Eq. (\ref{eqn:self-attention}) and $Q,K,V \in \mathbb{R}^{S \times T \times E}$. If $D=S$, then $Q_i,K_i,V_i \in \mathbb{R}^{T \times E}$, otherwise $Q_i,K_i,V_i \in \mathbb{R}^{S \times E}$. In practical implementations, we can perform Eq. (\ref{eqn:separable-attention}) efficiently by treating the non-aggregated dimension as an extra batch dimension. Figure. \ref{fig:train} shows an example of separable attention during training, where the \emph{target attention} denotes separable attention on $T$ and \emph{source attention} denotes separable attention on $S$.

  Similar to the masked self-attention in Transformer, a future mask is adopted to mask out the illegal softmax inputs in the target attention. This mask, as well as the parallelism among positions, allows the previously generated representations to be reused for computing only the sized $S \times E$ representation which is associated to the latest target tokens at each decoding step. We exemplify it in Figure. \ref{fig:decode}.

  \subsection{Reformer-base}

  Having the initial joint representation and separable attention, we present \emph{Reformer-base}. As shown in Figure. \ref{fig:reformer-base}, Reformer-base is simply a stack of layers. After computing the input $x \in \mathbb{R}^{S \times T \times E}$ according to Eq. (\ref{eqn:embed}), it will pass through four sublayers in each layer: the first one is the target attention that aggregates information along the dimension $T$, the second one is a point-wise feed-forward network as in Eq. (\ref{eqn:ffn}), the third one is the source attention that aggregates information along the dimension $S$ and the final one is another feed-forward network. All these sublayers are wrapped by the residual connection similar to Eq. (\ref{eqn:residual}). A graphical illustration of these sublayers is in Figure. \ref{fig:sublayers}.

  In the last layer, the network will produce the output with a size of $S \times T \times E$.  If a linear projection as well as a softmax is immediately followed, it will result in $S$ distinct predictions for each target position, while we expect only one. Thereby a \emph{Reduction} component is introduced to transform an input $x \in \mathbb{R}^{S \times E}$ to an output $y \in \mathbb{R}^{E}$ before the linear projection and softmax so as to obtain one prediction for one target position. The reduction is defined as:
  \begin{equation}
    \begin{split}
      \mathrm{Reduction}(x)&=\mathrm{Concat}(\mathrm{head}_1,\cdots,\mathrm{head}_E)\\
      \mathrm{where}\ \mathrm{head}_i&=\mathrm{softmax}(W_ix^T)x_i
    \end{split}
    \label{eqn:reduction}
  \end{equation}
  where $W$ are the learnable parameters and $W_i \in \mathbb{R}^{1 \times E}$ and $x_i \in \mathbb{R}^{S \times 1}$. The input and output of the reduction will go through a layer normalization first before any subsequent processing. This simple attention-like mechanism uses all $E$ features of a token to determine the importance of a single feature and normalizes it with other $S$ features in the same position but from different tokens. The rightmost part of Figure. \ref{fig:sublayers} demonstrates how reduction runs.

  Another important detail of Reformer-base is Dropout. From the initial construction of the joint representation in Eq. (\ref{eqn:embed}), we see that the information of a source token is distributed along the dimension $T$ and for the target token along the dimension $S$. This means that Dropout is no longer able to regularize model by encouraging feature independency within a token representation, as the model can access to the dropped features from representations of other tokens. We introduce Dropout1d and Dropout2d to alleviate this problem \cite{cvpr2015:Tompson}. In principle, the standard dropout noise is sampled independently for each feature. To prevent the feature dependency along specific dimensions, sharing dropout noise in these dimensions will mask out all potential duplicates that might creep into the model. We apply Dropout2d to the $S$ and $T$ dimensions of the output of the feed-forward network and Dropout1d to the $S$/$T$ dimension of the output of the target/source attention. We do so because for the feed-forward network it operates on the whole $\mathbb{R}^{S \times T \times E}$ space and we need to prevent feeding dropped features back via $S$ and $T$ dimensions, while for the separable attention it aggregates representations along one temporal dimension thus we only block feature dependency in the other one so as to not affect the aggregation.

  \begin{table}[t!]
    \centering
    \begin{tabular}{lcc}
      \toprule
      Layer Type & Complexity & Path Length\\
      \midrule
      Separable Attention & $O(n^3d)$ & $O(1)$\\
      Self-Attention & $O(n^2d)$ & $O(l)$\\
      Recurrent & $O(nd^2)$ & $O(l+n)$\\
      Convolution & $O(knd^2)$ & $O(l+n\log_k(n))$\\
      \bottomrule
    \end{tabular}
    \caption{The path length required to access any source or target token, $n$ is the sequence length, $d$ is the representation size, $k$ is the kernel size of convolutions and $l$ is the number of layers.}
    \label{tab:theory}
  \end{table}

  \section{Trading-Off the Effectiveness \& Efficiency }

  \subsection{The Pros \& Cons of Reformer-base}

  As noted in Table \ref{tab:theory}, Reformer-base can access the information of any source or target token in a path with the minimum $O(1)$ operations. Transformer with the self-attention layer requires $O(l)$ operations as the source side information is only available after passing through $l$ layers in the encoder. The same also happens in other layer types, resulting in the $l$ occurs inside their path lengths. This observation implies that Reformer-base can better capture the dependencies between the source and target tokens, because the path length for propagating the source side information is minimized. Although the separable attention of Reformer-base has higher complexity, it also has a higher degree of parallelism, as it computes all $S \times T$ positions simultaneously.

  Despite the theoretical effectiveness of Reformer-base, it suffers from two important issues that hinder its efficiency:
  \begin{description}
    \item[Duplicate Computation] When a new target token is fed into the model during decoding, its inner representation is computed only based on the output of the previous layer and starts from the embedding, as depicted in Figure. \ref{fig:decode}. This means that the high-level information about the source sentence and previously generated target tokens is unavailable for the lower layers since it is usually stored in the higher layers representations. In each decoding step, the model will have to recompute this information from scratch, which wastes the model capacity.
    \item[Computation Allocation] Another less obvious problem of Reformer-base is that it assigns the same amount of computation to both the source and target side, as in each layer the model has one source attention on $S$ and one target attention on $T$. It is undesirable since the number of source tokens fed into the model is significantly more than the one of target tokens at each decoding step. This observation implies that there should be more computation allocated to process the source side input \cite{acl2019:Wang}. Reformer-base can only stack more separable attention for compensation, where the redundant target attention introduces extra computation cost as noted by the separable attention complexity in Table. \ref{tab:theory}.
  \end{description}

  \subsection{Reformer-fast}

  Directly feeding the higher layer output from the preceding decoding steps into the lower layer input might resolve these two issues of Reformer-base. But this naive approach is not feasible as its sequential nature breaks the training parallelism by forcing the computation of the representation in one layer and one decoding step to wait for all representation below or before it finished computing, giving the total $O(LT)$ training cost, where $L$ is the total number of layers and $T$ is the target sentence length. Yet for Reformer-base, each of its layers only relies on the previous layer output, thus $O(LT)$ training cost can reduce to $O(L)$ by paralleling the computation over different steps.

  The solution we adopted here is that before computing the joint representation in Eq. (\ref{eqn:embed}), we introduce a preprocessing network \emph{PreNet} to process the source token embedding $\mathrm{embed}_i$ first and the network output is used to replace $\mathrm{embed}_i$ in Eq. (\ref{eqn:embed}). This network provides not only as much extra computation as required for the source tokens, but also a reusable high-level abstraction of these tokens. It consists of a stack of layers and a layer normalization at the end, where each layer is composed of a self-attention in Eq. (\ref{eqn:self-attention}) and a feed-forward network in Eq. (\ref{eqn:ffn}) with both wrapped as in Eq. (\ref{eqn:residual}). PreNet shares the same hyper-parameters setting as Reformer-base, e.g., the hidden size of the feed-forward network. PreNet is very efficient, as it is computed only once and reused at each decoding step. Figure. \ref{fig:preprocess} shows the PreNet architecture. Reformer-base together with the PreNet forms our new design, \emph{Reformer-fast}.

  Reformer-fast has higher efficiency than Reformer-base, cause it avoids stacking many high complexity separable attentions by reusing the high-level abstraction provided in the low complexity PreNet, i.e., from $O(n^3d)$ to $O(n^2d)$. But Reformer-fast scarifies the effectiveness of Reformer-base since it can no longer access any source or target token with $O(1)$ operation but $O(l)$ as the self-attention counterpart.

  \subsubsection{Discussion}

  Reformer-fast introduces one additional hyper-parameter to be tuned, i.e., the number of layers of PreNet, which denotes how much extra computation is allocated to process the source sentence. Assuming the source sentence length $N$ equals to the target sentence length, there will be $N$ source tokens and $t$ target tokens are fed into the model at the $t$-th decoding step. This implies that total $N^2$ source tokens and $\frac{1}{2}N^2+\frac{1}{2}N$ target tokens are presented in the whole decoding process. Considering only the second-order terms, the ratio of source tokens count to target tokens count $N^2/\frac{1}{2}N^2=2$ gives the insight that the computation allocated to the source side should be roughly $2\times$ more than the one of the target side. Therefore, we set the PreNet to have the same height as Reformer-base.
  
  \section{Toward Optimal Model Scaling}

  \subsection{Understanding Reformer}

  To scale Reformer to datasets that are larger than the one that the base setup is optimized for, we need to better understand both the network capacity and parameters. Unfolding the residual connection of the input $x_L$ in the last linear classifier layer, we have:
  \begin{equation}
    \begin{split}
      x_L&=x_{L-1}+f(x_{L-1})\\
      &=x_0+f(x_0)+\cdots+f(x_{L-1})
    \end{split}
    \label{eqn:unfold}
  \end{equation}
  where $x_0$ stands for the embedding, $x$ is the layer input and $f$ is the layer. Three factors that determine the network capacity are observed: the number of layers $l$, the embedding size $e$ and the hidden size of the feed-forward network $h=w \times e$, where larger $l$ and $e$ provide more features for the last layer and higher $w$ strengthens the capacity of each layer thus the output features quality.

  $l,w,e$ are also the dominant factors of network parameters. Omitting the bias terms and layer normalization, we have the number of parameters $P(l,w,e)=2le^2(4+2w)$, where we do not consider the embedding matrix here as the involved embedding are negligible on-the-fly. We can observe that $P$ grows quadratically with $e$, thus enlarging $e$ will overwhelm the contributions of $l,w$ and expose the model to the high risk of overfitting. Therefore we leave $e$ out of consideration for either scaling or calculating parameters. This gives a simplified formulation of parameters:
  \begin{equation}
    P(l,w)=2l(4+2w)
    \label{eqn:params}
  \end{equation}
  Then the problem of scaling becomes finding $l$ and $w$ that maximize the model performance on the validation set:
  \begin{equation}
    \max\limits_{l,w} \mathcal{L}(l,w)
    \label{eqn:scaling}
  \end{equation}
  where $\mathcal{L}$ is any performance measurement metric. In our experiments, we use the cross-entropy loss instead of BLEU as $\mathcal{L}$ because it is more stable.

  \subsubsection{Discussion}
  
  Eq. (\ref{eqn:unfold}) reveals that the superiority of Reformer comes from the fact that it has a larger input space thus richer input features, i.e., from $\mathbb{R}^{l \times e \times S}$ of the encoder or $\mathbb{R}^{l \times e \times (T + S)}$ of the decoder to $\mathbb{R}^{l \times e \times S \times T}$ of Reformer, where $S$ is the source sentence length and $T$ is the target sentence length. This also suggests that Reformer is a special scaled Transformer that scales $e$ dynamically.

  \subsection{Single-Shot Gradient-Based Scaling}

  Though many research have attempted to find the concrete formulation between the generalization and model complexity \cite{iclr2017:Zhang}, there is still no clear conclusion yet. Hence we can only rely on the black-box optimization algorithms to solve Eq. (\ref{eqn:scaling}), e.g., Grid Search.

  However, most black-box optimization algorithms are extremely inefficient in our case despite their superior empirical performances, because they require to evaluate $\mathcal{L}$ with different $l,w$ many times while a single evaluation is a complete training process. An efficient way is to compute the approximate gradient of $\mathcal{L}$ with respect to the current $l,w$ and perform only one step gradient ascent along with a step size $\alpha$ to scale $l,w$. By the infinitesimal definition of the partial derivative, we have the gradient of $\mathcal{L}$ w.r.t. $l$:
  \begin{equation}
    \mathcal{L}'_l = \lim_{\delta\to0}\frac{\mathcal{L}(l+\delta,w)-\mathcal{L}(l,w)}{\delta} \approx \frac{\mathcal{L}(l+\epsilon,w)-\mathcal{L}(l,w)}{\epsilon}
  \end{equation}
  where $\epsilon$ is a small number that we manually pick for the approximation. The same method is also adopted for retrieving $\mathcal{L}'_w$. Although the approximated gradients are only useable within a small region around the current $l,w$ since it is only a local direction indicator, it is sufficient when the step size $\alpha$ is small as we need not deviate from the base setup much.

  \begin{table*}[t!]
    \centering
    \begin{tabular}{l|c|c|c|c|c|c|c|c|c}
      \hline
      \multicolumn{1}{c|}{\multirow{2}*{System}} & \multicolumn{2}{c}{Vi-En} & \multicolumn{2}{|c}{De-En} & \multicolumn{2}{|c}{En-De} & \multicolumn{3}{|c}{Zh-En} \\
      \cline{2-10}
      & tst2012 & tst2013 & valid & test & valid & test & MT06 & MT05 & MT08 \\
      \hline
      baseline & 24.70 & 27.53 & 34.44 & 33.63 & 28.19 & 27.54 & 49.63 & 48.23 & 43.10 \\
      \hline
      Reformer-base & 24.42 & 27.18 & \textbf{35.87} & \textbf{34.92} & \textbf{29.42} & 28.32 & 50.00 & 48.72 & \textbf{45.04} \\
      Reformer-fast & \textbf{24.98} & \textbf{28.26} & \textbf{35.87} & 34.87 & 29.31 & \textbf{28.36} & \textbf{50.82} & \textbf{49.29} & 44.64 \\
      \hline
    \end{tabular}
    \caption{BLEU results of NMT systems.}
    \label{tab:bleu}
  \end{table*}

  The next problem is to choose a proper step size $\alpha$. When scaling up a network, we would like to maximize its performance while its risk of overfitting as well as the resource requirement is under controlled. In most cases, the number of parameters $P$ expresses the risk of overfitting, e.g., the use of $L_0$-Norm regularization term, and the required resources such as space and time also grow with $P$. Therefore the resource requirement and the risk of overfitting can be jointly represented by the parameters. Choosing a step size $\alpha$ under the parameters constraints is then equivalent to solving:
  \begin{equation}
    \frac{P(l+\alpha\mathcal{L}'_l,w+\alpha\mathcal{L}'_w)}{P(l,w)} \approx \beta
    \label{eqn:constraint}
  \end{equation}
  where the manually given constraint $\beta$ is the ratio we want to scale current parameters, i.e., having $\beta \times$ parameters after scaling. Solving Eq. (\ref{eqn:constraint}) with $P$ replaced by Eq. (\ref{eqn:params}) will produce a new promising $l$ and $w$ setup, denoted as $\hat{l}=l+\alpha\mathcal{L}'_l$ and $\hat{w}=w+\alpha\mathcal{L}'_w$ respectively.

  \subsubsection{Discussion}
  
  Compared to other hyper-parameter tuning algorithms \cite{2017:Golovin}, our approach is not directly applicable to hyper-parameters except $l,w$, because they do not contribute to Eq. (\ref{eqn:constraint}), therefore finding the optimal step size for them reduces to the Grid Search.

  \begin{table}[t!]
    \centering
    \begin{tabular}{lcccc}
      \toprule
      \multicolumn{1}{c}{System} & PPL & BLEU & Params & Speed \\
      \midrule
      baseline & 5.39 & 34.44 & 16M & $1\times$ \\
      Reformer & 5.00 & 35.16 & 16M & $0.47\times$ \\
      +Dropout 1/2d & 4.82 & 35.87 & 16M & $0.52\times$ \\
      +PreNet & 4.89 & 35.87 & 17M & $0.73\times$ \\
      \bottomrule
    \end{tabular}
    \caption{Ablation study on De-En validation set.}
    \label{tab:ablation}
  \end{table}

  \section{Experiments}

  \subsection{Setup}

  \subsubsection{Dataset}

  We evaluated our approach on IWSLT15 Vietnamese-English (Vi-En), IWSLT14 German-English (De-En), English-German (En-De) and NIST12 Chinese-English (Zh-En) translation tasks. For Vi-En translation, the training set consisted of 130K sentence pairs and we used tst2012 as the validation set and tst2013 as the test set. For De-En and En-De translations, the training set consisted of 160K sentence pairs and we randomly drew 7K samples from the training set as the validation set. We concatenated dev2010, dev2012, tst2010, tst2011 and tst2012 as the test set. For Zh-En translation, We used 1.8M sentence Chinese-English bitext provided within NIST12 OpenMT\footnote{LDC2000T46, LDC2000T47, LDC2000T50, LDC2003E14, LDC2005T10, LDC2002E18, LDC2007T09, LDC2004T08}. We chose the evaluation data of NIST MT06 as the validation set, and MT05, MT08 as the test set. All Chinese sentences were word segmented by a language model-based toolkit.

  \subsubsection{Model}

  Our baseline systems were based on the open-source implementation of the Transformer model \cite{nips2017:Vaswani} presented in \citet{naacl2019:Ott}. The model consisted of the 6-layer encoder and decoder. The size of the embedding, the heads and hidden layer of the feed-forward network were set to 256/512, 4/8 and 1024/2048 for the IWSLT/NIST datasets. Dropout was set to 0.1 for all experiments. For training, we used the Adam optimizer \cite{iclr2015:Kingma} where the learning rate and batch size were set to 0.0007 and 4096$\times$8 tokens. We applied BPE \cite{acl2016:Sennrich} to the De-En, En-De and Zh-En tasks except for the Vi-En one, where the word-based vocabulary achieved better performance.
  
  For both Reformer-base and Reformer-fast, we adopted the same settings as the Transformer baseline, except that Reformer-base consisted of 7 layers and for Reformer-fast it had 5 layers so as to obtain the similar number of parameters as the baseline. In the scaling experiments, we doubled the embedding size and set dropout to 0.3 for Transformer as provided in \citet{naacl2019:Ott} for IWSLT14 De-En translation and used Transformer-big setting for NIST12 Zh-En translation. As for Reformer-fast, we used $\beta=2$ to constraint the added parameters. All experiments were done on 8 Titan V GPUs with the half-precision training.

  \begin{table}[t!]
    \centering
    \begin{tabular}{l|c|c|c|c}
      \hline
      \multicolumn{1}{c|}{\multirow{2}*{System}} & \multicolumn{2}{c|}{De-En} & \multicolumn{2}{c}{Zh-En}\\
      \cline{2-5}
      & test & Params & MT08 & Params\\
      \hline
      baseline & 33.63 & 16M & 43.10 & 101M \\
      +scaling & 34.41 & 42M & 44.60 & 291M \\
      \hline
      Reformer-fast & 34.87 & 17M & 44.64 & 105M \\
      +scaling & \textbf{35.11} & 27M & \textbf{46.66} & 146M \\
      \hline
    \end{tabular}
    \caption{BLEU results of scaled systems.}
    \label{tab:scale}
  \end{table}

  \subsection{Results}

  As shown in Table \ref{tab:bleu}, both Reformer-base and Reformer-fast significantly outperform the Transformer baseline by a considerable margin in almost all test sets. For Vi-En translation, Reformer-base slightly underperforms the baseline. We attribute this to the symmetric computation allocation for both the source and target side in Reformer-base, as the Vietnamese sentences are typically longer than their English translations thus requesting more source-side operations. For De-En and En-De translations, Reformer-base outperforms Reformer-fast in most cases, where the symmetric computation allocation property better suits these datasets since German sentences are often of similar lengths as their English equivalents. As for Zh-En translation, none of the Reformer models outperforms the other one in all test sets, which indicates that both models behave similarly well.
  
  We perform the ablation study of Reformer models on the IWSLT14 De-En validation set. According to Table \ref{tab:ablation}, we find that dropout 1/2d did better regularize our model with 0.7 BLEU point improvement. With the same amount of parameters, adding PreNet achieves comparable performance but nearly 50\% faster than the one without PreNet in decoding, justifying that Reformer-fast did better trade-off the effectiveness and efficiency than Reformer-base.
  
  In the scaling experiments, we obtain $\mathcal{L}'_l=0.01063$ and $\mathcal{L}'_w=0.01069$ by picking $\epsilon=1$, which implies that depth and width are of similar importance to Reformer-fast. Solving Eq. (\ref{eqn:constraint}) under $\beta=2$ gives the approximate solution of adding 2 layers and $1/2$ hidden size. Table \ref{tab:scale} shows the test set performances after applying the resulting scaling strategy. We observe that our scaling approach indeed generates a stronger model with better generalization ability, which outperforms the SOTA baseline by 0.7/2 BLEU points in the De-En/Zh-En translation with only $\sim$50\% parameters.
  
  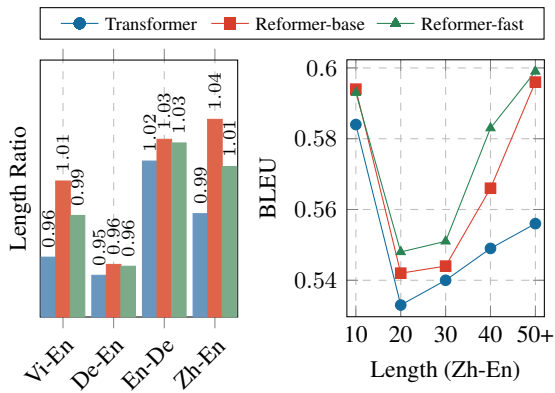
\begin{figure}[t!]
    \centering
    \begin{tabular}{@{\hskip0pt}r@{\hskip0pt}}
      \tikz {
        \scriptsize
        \node () at (6.68,0) {};
        \node (l0) at (-0.15,0) {};
        \node (l4) at (6.3,0) {};
        \draw[myblue] (0,0) -- plot[mark=*](0.25,0) -- (0.5,0) node[black,right] (l1) {Transformer};
        \draw[myred] (2,0) -- plot[mark=square*](2.25,0) -- (2.5,0) node[black,right] (l2) {Reformer-base};
        \draw[mygreen] (4.2,0) -- plot[mark=triangle*](4.45,0) -- (4.7,0) node[black,right] (l3) {Reformer-fast};
        \begin{pgfonlayer}{background}
          \node[rectangle,draw,inner sep=1pt] [fit = (l0) (l1) (l2) (l3) (l4)] {};
        \end{pgfonlayer}
      }\\[0pt]
      \hspace*{\fill}
      \begin{tikzpicture}
        \begin{axis}[
          width=0.5\linewidth,height=5cm,
          symbolic x coords={Vi-En,De-En,En-De,Zh-En},
          enlarge x limits=0.15,
          enlarge y limits={upper,value=0.3},
          ylabel={Length Ratio},
          ylabel near ticks,
          ybar=0pt,
          xtick=data,
          ytick=\empty,
          ymin=0.93,
          nodes near coords,
          every node near coord/.append style={rotate=90,anchor=west,font=\scriptsize},
          every tick label/.append style={font=\footnotesize},
          xticklabel style={rotate=45,anchor=north east,font=\footnotesize,inner sep=0pt,outer sep=2pt},
          bar width=5.5pt,
          xmajorgrids=true,
          ymajorgrids=true,
          grid style=dashed,
          label style={font=\small},
        ]
          \addplot [draw=myblue!60,fill=myblue!60] coordinates {(Vi-En,0.963) (De-En,0.953) (En-De,1.016) (Zh-En,0.987)};
          \addplot [draw=myred!80,fill=myred!80] coordinates {(Vi-En,1.005) (De-En,0.959) (En-De,1.028) (Zh-En,1.039)};
          \addplot [draw=mygreen!60,fill=mygreen!60] coordinates {(Vi-En,0.986) (De-En,0.958) (En-De,1.026) (Zh-En,1.013)};
        \end{axis}
      \end{tikzpicture}
      \hfill
      \begin{tikzpicture}
        \begin{axis}[
          width=0.5\linewidth,height=5cm,
          enlargelimits=0.05,
          ylabel={BLEU},
          ylabel near ticks,
          xlabel={Length (Zh-En)},
          xlabel near ticks,
          symbolic x coords={10,20,30,40,50+},
          xmajorgrids=true,
          ymajorgrids=true,
          grid style=dashed,
          xtick=data,
          legend style={font=\scriptsize,at={(0.5,1.02)},anchor=south,legend columns=3,/tikz/every even column/.append style={column sep=7pt}},
          every tick label/.append style={font=\small},
          label style={font=\small},
        ]
          \addplot [myblue,mark=*] coordinates {
            (10,0.584) (20,0.533) (30,0.540) (40,0.549) (50+,0.556)
          };
  
          \addplot [myred,mark=square*] coordinates {
            (10,0.594) (20,0.542) (30,0.544) (40,0.566) (50+,0.596)
          };
  
          \addplot [mygreen,mark=triangle*] coordinates {
            (10,0.593) (20,0.548) (30,0.551) (40,0.583) (50+,0.599)
          };
        \end{axis}
      \end{tikzpicture}
      \hspace*{\fill}\\[0pt]
    \end{tabular}
    \caption{Length statistics.}
    \label{fig:length}
  \end{figure}

  \subsection{Analysis}

  We investigate the effectiveness of the proposed two Reformer variants. The left of Figure. \ref{fig:length} demonstrates the ratio of the translations lengths to the references lengths in various systems. We observe that Reformer models generate translations with more appropriate lengths than the baseline, i.e., less likely to have under-translation. Interestingly, Reformer-base has higher length ratio than the others, implying it tends to generate longer translations. The right of Figure. \ref{fig:length} shows the BLEU score of translations under source sentences of different lengths. Reformer models outperform the baseline the most in long sentences, which implies that they better capture the long-term dependencies.

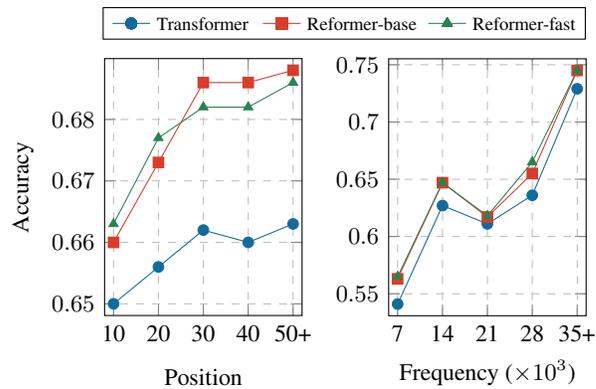
\begin{figure}[t!]
    \centering
    \begin{tabular}{@{\hskip0pt}r@{\hskip0pt}}
      \tikz {
        \scriptsize
        \node () at (6.55,0) {};
        \node (l0) at (0,0) {};
        \node (l4) at (6,0) {};
        \draw[myblue] (0,0) -- plot[mark=*](0.25,0) -- (0.5,0) node[black,right] (l1) {Transformer};
        \draw[myred] (2,0) -- plot[mark=square*](2.25,0) -- (2.5,0) node[black,right] (l2) {Reformer-base};
        \draw[mygreen] (4.2,0) -- plot[mark=triangle*](4.45,0) -- (4.7,0) node[black,right] (l3) {Reformer-fast};
        \begin{pgfonlayer}{background}
          \node[rectangle,draw,inner sep=1pt] [fit = (l0) (l1) (l2) (l3) (l4)] {};
        \end{pgfonlayer}
      }\\[0pt]
      \hspace*{\fill}
      \begin{tikzpicture}
        \begin{axis}[
          width=0.5\linewidth,height=5cm,
          enlargelimits=0.05,
          ylabel={Accuracy},
          ylabel near ticks,
          xlabel={Position},
          xlabel near ticks,
          symbolic x coords={10,20,30,40,50+},
          xmajorgrids=true,
          ymajorgrids=true,
          grid style=dashed,
          xtick=data,
          every tick label/.append style={font=\small},
          xlabel style={yshift=-0.125cm},
          label style={font=\small},
        ]
        \addplot [myblue,mark=*] coordinates {
          (10,0.650) (20,0.656) (30,0.662) (40,0.660) (50+,0.663)
        };

        \addplot [myred,mark=square*] coordinates {
          (10,0.660) (20,0.673) (30,0.686) (40,0.686) (50+,0.688)
        };

        \addplot [mygreen,mark=triangle*] coordinates {
          (10,0.663) (20,0.677) (30,0.682) (40,0.682) (50+,0.686)
        };
        \end{axis}
      \end{tikzpicture}
      \hfill
      \begin{tikzpicture}
        \begin{axis}[
          width=0.5\linewidth,height=5cm,
          xlabel={Frequency ($\times 10^3$)},
          xlabel near ticks,
          enlargelimits=0.05,
          symbolic x coords={7,14,21,28,35+},
          xmajorgrids=true,
          ymajorgrids=true,
          grid style=dashed,
          xtick=data,
          every tick label/.append style={font=\small},
          label style={font=\small},
        ]
        \addplot [myblue,mark=*] coordinates {
          (7,0.541) (14,0.627) (21,0.611) (28,0.636) (35+,0.729)
        };

        \addplot [myred,mark=square*] coordinates {
          (7,0.563) (14,0.647) (21,0.617) (28,0.655) (35+,0.745)
        };

        \addplot [mygreen,mark=triangle*] coordinates {
          (7,0.565) (14,0.647) (21,0.618) (28,0.665) (35+,0.745)
        };
        \end{axis}
      \end{tikzpicture}
      \hspace*{\fill}\\[0pt]
    \end{tabular}
    \caption{Accuracy statistics (Zh-En).}
    \label{fig:acc}
  \end{figure}

  In addition to the length perspective, we provide a series of accuracy statistics for the analysis. According to the left of Figure. \ref{fig:acc}, we find that Reformer models have higher prediction accuracy than the baseline over all target positions within sentences, especially for the distant ones. This observation is on par with the one in the right of Figure. \ref{fig:length} that Reformer outperforms Transformer the most in long sequences. The right of Figure. \ref{fig:acc} presents how systems behave on target tokens of different occurrence frequencies. An interesting phenomenon is that Reformer predicts low-frequency tokens more accurate than Transformer, which means that it exploits the context better to infer rare tokens.

\begin{figure}[t!]
    \centering
    \hspace*{\fill}
    \begin{tikzpicture}
      \pgfplotstableread[col sep=space,string type]{
        word plot1 plot2 plot3 plot4 plot5
        w\o3 0.3079 0.1196 0.1074 0.1959 0.1981
        h\en3 0.1270 0.1080 0.2315 0.2089 0.0842
        h\ao3 0.1288 0.1885 0.3216 0.0633 0.1185
        . 0.2979 0.3339 0.2617 0.4556 0.3476
        EOS 0.1384 0.2499 0.0778 0.0763 0.2516
      }\tabledata
      \begin{axis}[
        width=0.5\linewidth,height=4cm,
        title={Source attention},
        zmin=0, zmax=0.4,
        area plot/.style={
          fill opacity=0.75,
          draw=orange!80!black,thick,
          fill=orange,
          mark=none,
        },
        xtick={0,...,5},
        ytick={1,...,5},
        xticklabels from table={\tabledata}{word},
        yticklabels={EOS,I,am,fine,.},
        xticklabel style={rotate=45,anchor=east,font=\scriptsize,inner sep=0pt,outer sep=0pt},
        yticklabel style={rotate=-45,anchor=west,font=\scriptsize,inner sep=0pt,outer sep=0pt},
        zticklabel style={rotate=90,anchor=south,font=\scriptsize},
        xticklabel shift={3pt},
        yticklabel shift={3pt},
        xmajorgrids=true,
        ymajorgrids=true,
        zmajorgrids=true,
        grid style=dashed,
      ]
        \pgfplotsinvokeforeach{5,4,...,1}{
          \addplot3 [area plot] table [
            x expr=\coordindex, y expr=#1, z=plot#1,
          ] {\tabledata} \closedcycle;
        }
      \end{axis}
    \end{tikzpicture}
    \hfill
    \begin{tikzpicture}
      \pgfplotstableread[col sep=space,string type]{
        word plot1 plot2 plot3 plot4 plot5
        EOS 0.1990 0.4948 0.4168 0.6119 0.4448
        I 0.4404 0.0745 0.0731 0.1582 0.0439
        am 0.2706 0.3310 0.3527 0.1884 0.4091
        fine 0.0900 0.0998 0.1575 0.0416 0.1022
        . 0.0000 0.0000 0.0000 0.0000 0.0000
      }\tabledata
      \begin{axis}[
        width=0.5\linewidth,height=4cm,
        title={Target attention},
        zmin=0, zmax=1,
        area plot/.style={
          fill opacity=0.75,
          draw=orange!80!black,thick,
          fill=orange,
          mark=none,
        },
        xtick={0,...,5},
        ytick={1,...,5},
        xticklabels from table={\tabledata}{word},
        yticklabels={w\o3,h\en3,h\ao3,.,EOS},
        xticklabel style={rotate=45,anchor=east,font=\scriptsize,inner sep=0pt,outer sep=0pt},
        yticklabel style={rotate=-45,anchor=west,font=\scriptsize,inner sep=0pt,outer sep=0pt},
        zticklabel style={rotate=90,anchor=south,font=\scriptsize},
        xticklabel shift={3pt},
        yticklabel shift={3pt},
        xmajorgrids=true,
        ymajorgrids=true,
        zmajorgrids=true,
        grid style=dashed,
      ]
        \pgfplotsinvokeforeach{5,4,...,1}{
          \addplot3 [area plot] table [
            x expr=\coordindex, y expr=#1, z=plot#1,
          ] {\tabledata} \closedcycle;
        }
      \end{axis}
      \node () at (0,-12pt) {};
    \end{tikzpicture}
    \hspace*{\fill}
    \caption{Attention (\emph{``w\o3 h\en3 h\ao3 \underline{.}''} $\to$ \emph{``I am \underline{fine} .''}).}
    \label{fig:attention}
  \end{figure}
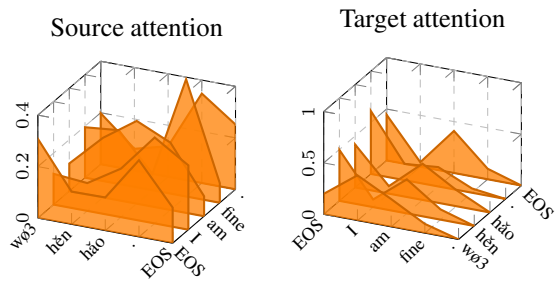

  Moreover, we illustrate an example of separable attention in Figure. \ref{fig:attention}. The left part is the source attentions of the source token \emph{``.''} to the other ones given different target tokens. These distributions vary when the target token changed, as opposed to a single fixed distribution in the Transformer encoder self-attention. The right part is the target attentions of \emph{``fine''} to previous target tokens given different source tokens. Each source token has its own distribution and it differs from each other, contrary to a single static distribution in the Transformer decoder self-attention. This suggests that Reformer passes more information through more channels.

  \section{Related Work}

  The problem of modelling the interaction in the machine translation systems has long been discussed for years. SMT systems explicitly represent the alignment \cite{book2009:koehn}, which directly account for one type of the interaction between any two source and target units. For NMT models, the interaction is only partially modelled via the attention, as it is restricted to one sequence rather than the cartesian product of two sequences. Recent works attempt to alleviate this issue through the joint representation. \citet{corr2015:Kalchbrenner} extends LSTM from one dimension to multi-dimensions and \citet{emnlp2018:Bahar} uses a two dimensions version to iterate through the joint representation. Perhaps the most related work is the pervasive attention model \cite{conll2018:Elbayad}. They first construct a joint representation from the source and target embedding, then stack several convolution layers to produce the final predictions. Our network is also built on top of the joint representation, but each layer can access to any source or target token in a path with the minimum $O(1)$ operation, compared to $O(n)$ for the 2D LSTM and $O(\log_k(n))$ for the pervasive attention, where $n$ is the sequence length and $k$ is the kernel size of the convolution.

  \section{Conclusion}

  We proposed two attention-based networks that use the joint representation to model the interaction. These models achieved significant improvements over four translation tasks, including the small and large scale ones. Despite their successes, we expect more future work on this type of models as they are still in a primitive form.

  \section{Acknowledgments}

  This work was supported in part by the National Science Foundation of China (Nos. 61876035, 61732005 and 61432013), the National Key R\&D Program of China (No. 2019QY1801) and the Opening Project of Beijing Key Laboratory of Internet Culture and Digital Dissemination Research. The authors would like to thank anonymous reviewers for their comments.

  \bibliography{reformer}
  \bibliographystyle{aaai}

\end{document}